\documentclass[12pt,journal,draftclsnofoot,onecolumn]{IEEEtran}
\usepackage{textcomp}
\usepackage{lineno}
\usepackage{graphicx}
\usepackage{amsmath}
\usepackage{balance}
\usepackage{multirow}
\usepackage{mathrsfs}
\usepackage{lscape}
\usepackage[tight,footnotesize]{subfigure}

\ifCLASSOPTIONcompsoc
  \usepackage[nocompress]{cite}
\else
  \usepackage{cite}
\fi

\begin{document}
\graphicspath{{./}}
\hyphenation{op-tical net-works semi-conduc-tor}

\author{Yu~Zhang,
        Stephen~Wistar,
        Jia~Li,
        Michael~Steinberg
        and~James~Z.~Wang
\IEEEcompsocitemizethanks{\IEEEcompsocthanksitem Yu Zhang and James Z. Wang are with the College
of Information Sciences and Technology, The Pennsylvania State University, University Park,
PA 16802, USA.\protect
\IEEEcompsocthanksitem Stephen Wistar and Michael Steinberg are with AccuWeather Inc.,
State College, Pennsylvania, USA\protect
\IEEEcompsocthanksitem Jia Li is with the Department of Statistics, Eberly College of Science,
The Pennsylvania State University, University Park, PA 16802, USA.
}
\thanks{Manuscript received --; revised --.}
}


\title{Storm Detection by Visual Learning Using Satellite Images}





\maketitle

\begin{abstract}
Computers are widely utilized in today's weather forecasting
as a powerful tool to leverage an enormous amount of data.
Yet, despite the availability of such data, current techniques 
often fall short of producing reliable detailed
storm forecasts. Each year severe thunderstorms
cause significant damage
and loss of life, some of which could be avoided
if better forecasts were available.
We propose a computer algorithm
that analyzes satellite images from historical archives
to locate visual signatures of 
severe thunderstorms for short-term predictions.
While computers are involved in
weather forecasts to solve numerical
models based on sensory data,
they are less competent in forecasting based on
visual patterns from satellite images.
In our system, we extract and summarize important visual storm
evidence from satellite image sequences in the way that meteorologists
interpret the images.
In particular, the algorithm extracts and fits local cloud motion
from image sequences to model the storm-related cloud patches.
Image data from the year 2008 have been
adopted to train the model, and historical thunderstorm
reports in continental US from 2000 through 2013 have been used as the ground-truth and priors
in the modeling process.
Experiments demonstrate the usefulness and potential of the algorithm
for producing more accurate thunderstorm forecasts.
\end{abstract}



\section{Introduction}\label{sec:intro}
\subsection{Background}

The numerical weather prediction (NWP) approach has been applied to weather
forecasting since the 1950s~\cite{lynch2008origins},
when the first computer ENIAC was invented~\cite{platzman1979eniac}.
John von Neumann, a pioneering computer scientist and meteorologist,
first applied the primitive equation models~\cite{charney1955use} on ENIAC.
Since then meteorologists have
produced reliable weather forecasts, and they have continued to
refine the numerical models, thereby improving the
NWP in recent decades. 
Today, the model of the European Centre for Medium-range Weather Forecasts (ECMWF)
and the United States' Global Forecast System (GFS) are two of the most accurate NWP models.
At the 500~hPa geopotential height (HGT) level of the atmosphere, both models
are reported to have annual mean correlation coefficients of nearly
0.9 for 5-day forecasts~\cite{yang2013review}. Other models,
such as UK Met Office Unified model~\cite{lean2008characteristics},
M\'{e}t\'{e}o-France ARPEGE-Climat model~\cite{deque1999documentation},
and Canada Meteorological Centre model~\cite{gauthier1999implementation},
are developed and play important roles in
global and regional weather predictions.

Even though NWP can effectively produce accurate weather
forecasts of the general weather pattern,
it is not always reliable in the prediction
of extreme weather events,
such as severe thunderstorms, hail storms,
hurricanes, and tornadoes, which
cause a significant amount of damage and loss of life
every year worldwide. Unreliable predictions,
either miss or false alarm ones, could
cause massive amounts of loss to the society.
The most well-known failed prediction is the Great Storm of 1987
in UK~\cite{burt1988great}.
The Met Office was criticized for not predicting
the storm timely. More recently, the landfall of 2012 Hurricane
Sandy in the northeastern US was not correctly predicted
by the National Hurricane Center until two days before it
came ashore~\cite{blake2013tropical}.
In January 2015, the impact of winter storm Juno
was overestimated by the US National Weather
Service~\cite{winkler2015importance}. Several cities
in US East Coast suffered from unnecessary travel bans and public
transportation closures. Much of the loss of life and property
due to improper precautions to severe weather events could be
avoided with more accurate (for both severity and location)
weather forecasts.

The numerical methods used in NWP can efficiently process large amounts of
meteorological measurements. However, a major drawback of such
methods is that they are sensitive to noise and initial inputs
due to the complexity of the models~\cite{lorenc20074d,doyle2014initial}.
As a result, although nowadays we have
powerful computers to run complex numerical models,
it is difficult to get accurate predictions computationally,
especially in the forecasts of severe weather.
To some extent, they
do not interpret the data from
a global point of view at a high cognitive level.
For instance, meteorologists can make good judgments of the
future weather conditions by looking at the general cloud patterns
and developing trends from a sequence of satellite
images by using geographical knowledge and their experience of past weather events~\cite{dvorak1975tropical,scofield1987nesdis};
numerical methods do not capture such high-level clues.
Additionally, historical weather records provide valuable
references for making weather forecasts, but numerical methods
do not make best use of them. 

To address this weakness of
numerical models, we develop a computational weather forecast
method that takes advantage of
both the global visual clues of satellite data
and the historical records. In particular, we
try to find synoptic scale features of mid-latitude  
thunderstorms (to be specified later in Section~\ref{sec:pattern})
by computer vision and machine learning approaches. It is worth mentioning 
that the purpose of this work is not to 
replace numerical models in forecasting. Instead, by tackling the
problem using different data sources and modeling methodologies, 
we aspire to develop a method
that can potentially be used side-by-side with and complement 
numerical models.

We analyze the satellite imagery because
it provides important clues as meteorologists view
evolving global weather systems.
Unlike conventional meteorological measures such as
temperature, humidity, air pressure, and wind speed,
which directly reflect the physical conditions
at the sensors' locations,
the visual information in satellite images
is captured remotely from
the orbit, which means a larger geographic
coverage with good resolution.
The brightness in an infrared satellite image indicates
the temperature of the cloud tops~\cite{liu2004satellite}
and therefore reveals
information about the top-level structure of the cloud systems.
Manually interpreting such visual information,
we can trace the evidence that corresponds to certain weather
patterns, particularly those related to severe thunderstorms. Moreover,
certain cloud patterns can be evident in the early stage of
storm development. As a result, the satellite imagery
are very helpful to meteorologists in their work.

Human eyes can effectively capture the visual patterns related to
different types of weather events.
The interpretation requires a high level understanding of
meteorological knowledge, which has been difficult for computers.
However, because the broad spatial and temporal coverage of the
satellite images produces too much information to be fully processed,
there is an emerging need for a computerized way to
analyze the satellite images automatically.
Therefore researchers are seeking computer vision algorithms to 
automatically process and interpret the visual features
from the satellite images, aiming to help meteorologists 
better analyze and predict weather events.
Traditional satellite image analysis techniques include
cloud detection with complex backgrounds~\cite{srivastava2003onboard},
cloud type classification~\cite{yang2004prep,behrangi2010day,price1990using},
detection of cloud overshooting top~\cite{bedka2010obje,bedka2011over},
and tracking of cyclone movement~\cite{ho2008auto}.
These approaches, however, only capture local visual features
without utilizing many high-level {\it global visual patterns},
and overlook their {\it development
over time}, which provides more clues for weather forecasts.
With the development of advanced computer vision algorithms,
some recent techniques put more focus on
the analysis of cloud motion and
deformation~\cite{zinner2008cb,keil2009displacement,merk2013detection,evans2006cloud}.
In these approaches, 
cloud motions are estimated to match and compare cloud patches
between adjacent images in a sequence.
In this paper, the proposed algorithm also uses cloud motion estimation
in image sequences. It is different from existing work 
in that it extracts and models certain patterns
of cloud motion, in addition to capturing the cloud displacement.


Historical satellite data archives as well as
meteorological observations are readily available
for recent years. By analyzing large amounts of
past satellite images and inspecting the historical
weather records, our system takes advantage of big
data to produce storm alerts given a certain query
image sequence. The result provides
an alternative prediction that can validate and correct
the conventional forecasts. It can be
embedded into an integrative or information fusion system,
as suggested in~\cite{mcgovern2006open},
to work with data and decisions from other sources
and produce more improved storm forecasts.

\subsection{The Dataset}\label{sec:prelim}

For our research, we acquired all of the GOES-M~\cite{noaa1998earth}
satellite imagery for the year of 2008\footnote{The
satellite imagery is publicly viewable and
the raw data archive can be requested
on the website of US National Oceanic and
Atmospheric Administration (NOAA).},
which was an active year containing
abundant severe thunderstorm cases for us to analyze.
The GOES-M satellite moves on a geostationary orbit and
was continually facing the North America area during that year.
In addition to the imagery data, each record contains
navigation information, which helps us map
each image pixel to its real geo-location.
The data is multi-spectral and contains five channels
covering different ranges of wavelengths, among which
we adopted channel 3 (6.5-7.0$\mu$m) and channel 4 (10.2-11.2$\mu$m)
in our analysis because these two infrared channels
are available both day and night. Both channels are of 4~km resolution
and the observation frequency is four records per hour,
i.e., about 15 minutes between the adjacent images.
To make sure two adjacent images in a sequence have a
noticeable difference, we sampled the original sequence
and used a sub-sequence with a 2-frames-per-hour frame
rate. We select the 30-minute sampling interval
because the algorithm is based on the optical flow analysis
between to adjacent image frames. We need both
evident enough visual difference and relatively
good temporal resolution. The 30-minute interval
is a empirical selection that considers both factors.
In the rest of the paper, every two adjacent images
in a sequence we refer to are 30 minutes away
without specifying.

Using the imagery and navigation data blocks in each raw data entry,
we reconstruct satellite images by an equirectangular
projection~\cite{snyder1987map}.
The pixel mapping from the raw imagery data to the map coordinate
is computed by the navigation data with transformation defined
by~\cite{noaa1998earth}. The map resolution is set to be 4~km (per pixel), which
best utilizes the information of the raw data. We reconstruct
the aligned images within the range
of 60\textdegree W to 124\textdegree W in longitude and
20\textdegree N to 52\textdegree N in latitude,
which covers the Continental US (CONUS) area.
Hereafter
all the image related computations in our approach
are performed on the reconstructed images,
on which the geo-location of each pixel is directly accessible.

We studied the relationship between satellite images and
severe thunderstorms, which may include hailstorms and
tornados\footnote{Severe thunderstorms mostly contain
hailstorms and tornados. Storms with hail and/or tornadoes are 
a subset of thunderstorms. Hereafter we generally refer
to severe thunderstorms as storms.}.
To relate the satellite data with severe thunderstorms,
we retrieved the storm report
data from the NOAA National Weather
Service\footnote{http://www.spc.noaa.gov/climo/online/},
where all the time and locations of storms
inside the United States since the year of 2000 are archived.
The records act as ground-truth data in the training
and provide geographic and temporal priors for the
system to make decisions.

\subsection{Synoptic-scale Visual Storm Signatures}\label{sec:pattern}

Being evidently visible on the satellite imagery,
synoptic-scale (spanning several hundred kilometers) storm features 
are studied in this paper.
Particularly, we are interested in 
the synoptic-scale features of
the mid-latitude ($30^{\circ}$-$60^{\circ}$ in latitude)
thunderstorms located outside of the Hadley cell~\cite{marshall1965atmosphere}
of the atmospheric circulation
because storms cooccurring with these features
are one of the major components
of storms in the CONUS and other
densely populated areas around the world.

\begin{figure}[t]
\centerline{\includegraphics[width=0.8\textwidth]{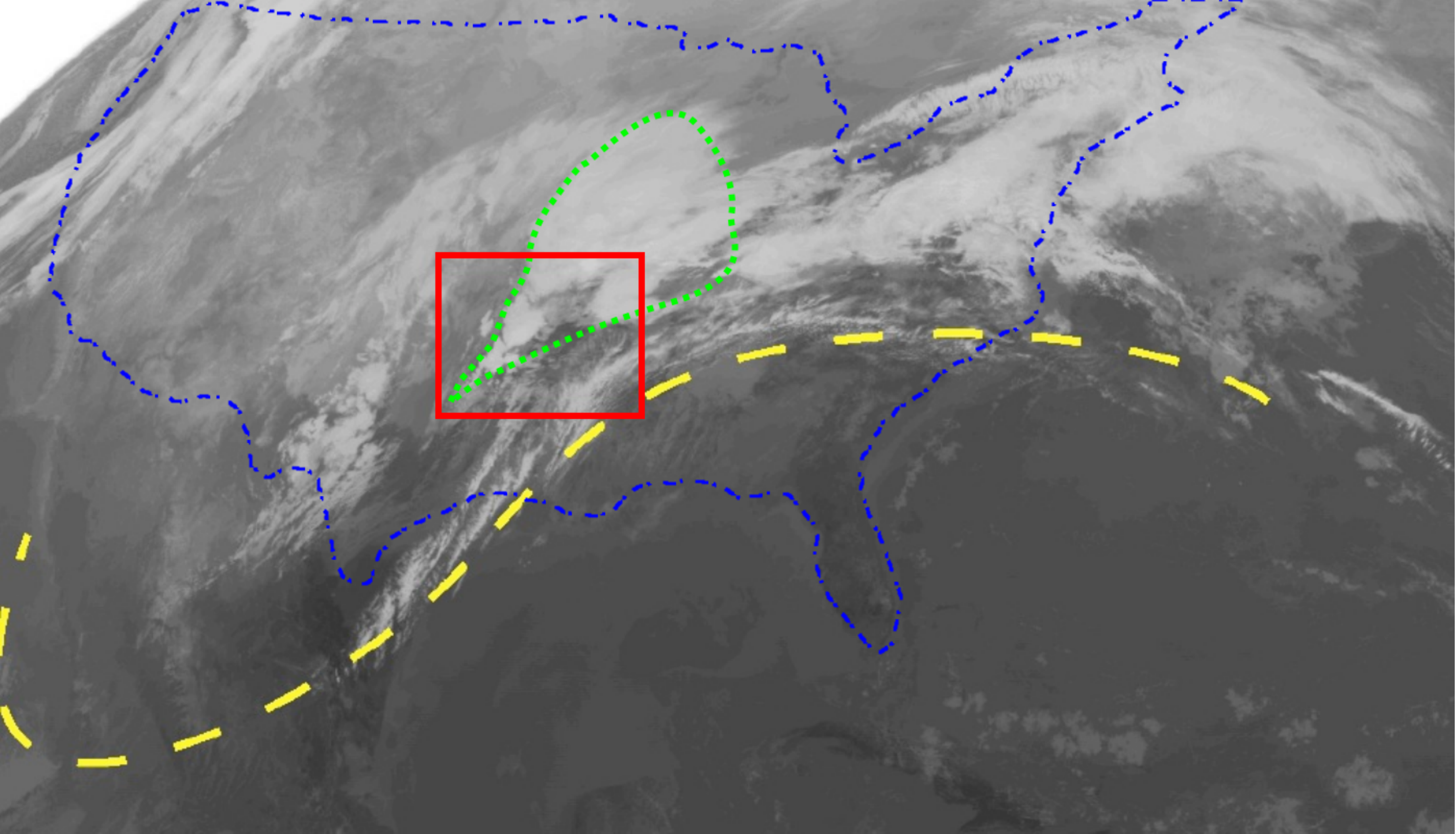}}
\caption{A sample GOES-M satellite channel 4 image taken at
16:02~(GMT) on February 5, 2008. 
The boxed area underwent a severe
thunderstorm later on that day.
The jet stream (marked in dashed curve) can be
implied from the distribution of clouds.
The storm area is covered by a comma-shaped
cloud patch (surrounded by dashed boundary)
in the trough region.} \label{fig:traugh}
\vspace{-0.1in}
\end{figure}

Synoptically, one important atmospheric feature
helpful for locating thunderstorms is the subtropical
{\it jet stream}\footnote{Hereafter we refer to the subtropical jet stream as the jet stream for
simplicity, regardless of the existence of the polar jet stream.},
which is the high-altitude (10-16~km) westerly wind
near the boundary between the Hadley cell
and the Ferrel cell; in other words, the cold and warm air masses.
Jet streams move in a meandering path from west to east,
a direction of movement caused
by the {\it Coriolis effect}~\cite{barry2009atmosphere}.
In the northern hemisphere, elongated cold air masses with
low pressure, namely ``troughs'', 
will have a dip in the jet stream extending southward at
various longitudes around the earth while in other locations
a ridge of higher pressure with warm air will send the jet
stream bulging northward.
The east side of a
trough, which has a southwest to northeast oriented jet stream, is a region of active
atmospheric storminess~\cite{clark2009climatology},
where cold and warm air masses collide.
Though jet streams themselves are invisible,
they can be revealed by thick cumulus clouds gathering in the
unsettled trough regions and areas of clouds spreading
northeastward along the east side of a trough. 
As a result, an elongated cloud-covered area along a southwest
to northeast direction is a useful large-scale clue for us to locate storms.
Fig.~\ref{fig:traugh}
shows a GOES-M channel-4 image taken on February 5, 2008.
Clearly in the mid-latitude region
the southern boundary of the cloud band
goes along a smooth curve, which 
indicates the jet stream
(sketched in dashed line).
Usually the cloud-covered areas to
the northwest side of such an easily-perceivable
jet stream are susceptible to storms.
In this case, severe thunderstorms developed two hours
after the time of the image in the boxed area of the
figure. Based on our observation, a large proportion of storms in
the CONUS area have similar cloud
appearances as in this example. 
As a result, finding jet streams in the satellite image
is important for storm prediction~\cite{charney1947dynamics}.

A jet stream travels along a large-scale meandering path,
whose wavelength is about
60$^{\circ}$ to 120$^{\circ}$ of longitude long,
as wide as a whole continent. 
The full wavelength is therefore usually referred to as a {\em long wave}.
The trough within a long wave
only tells us a general large region that is vulnerable
to storms. To locate individual storm cells more precisely,
meteorologists look for more 
detailed smaller scale cloud patterns caused by 
{\it short waves} traveling through the long wave.
Short waves are synoptic scale features with the wavelength
of the order of 1,000 km, and indicate the disturbance of
the mid or upper part of the atmosphere. Horizontally
seen from a satellite image, they appear as
smaller waves in the long wave trough.
In many cases, we can observe clouds in
``{\it comma shapes}''~\cite{carlson1980airflow,wallace2006atmospheric}
in short waves, i.e., such a cloud has
a round shape due to the
circular wind in its vicinity and a short tail
towards the opposite direction of its translation.
A comma-shaped cloud patch indicates a turbulent
area of rising air often leading to
convection (thunderstorms), especially
if it lies on the boundary between cold and warm
air masses.
Fig.~\ref{fig:traugh} shows an example of a comma-shaped cloud.
A storm area is marked by an square in the tail of the
comma shape in the figure.
Because of its distinctive pattern,
the comma shape has been one of the most utilized signature
when meteorologists review satellite images for thunderstorms.

It is necessary to mention
that the comma shape is still a signature in the
synoptic scale and therefore used for generally
locating a storm system rather than individual
storms\footnote{Individual thunderstorms typically
develop comma shapes to their clouds,
but not until they have become
e more mature. Such small-scale scale signatures
are also more difficult to detect in the satellite
image and therefore not the focus of this paper.}.
In other words, groupings of thunderstorms
tend to have a comma shape and the shape is used
as a first approximation of where severe
weather is most likely to occur.

The visual cloud patterns introduced above can often be
detected from a single satellite image.
However, a single image is not sufficient to provide
all the crucial information about the
dynamic of a storm system. The cloud shapes
are not always clear due to the irregular nature of clouds.
In addition, the evolution of storm systems, including
cloud areas  emerging, disappearing, and deforming,
cannot be revealed unless different satellite
images in a sequence are compared.

In practice, an important step in producing
weather forecasts is to review satellite images,
during which process meteorologists typically review 
a series of images, rather than a single one,
to capture critical patterns.
They use the differences between the key visual
clues (e.g., key points, edges, etc.) to track
the development of storm systems and better locate
the jet streams and comma shapes, among other features.
As mentioned in~\cite{doswell2004weather},
human heuristics are helpful in improving weather forecasts.
Following such guideline, our algorithm attempts to simulate
meteorologists' cognitive processes in analyzing temporally
adjacent satellite image frames.
In particular, two types of motions are regarded
crucial to the storm visual patterns:
the global cloud translation with the jet stream
and the local cloud rotations producing the comma shapes.
We consider both types of motions for extracting synoptic
storm signatures.


\subsection{Objectives}
The algorithm proposed in this paper aims at assisting the prediction
of severe thunderstorms from a different perspective from the NWP.
As an initial attempt in this direction, 
we focus on short-term severe thunderstorm prediction.
Synoptical scale features abstracted from known visual
storm patterns as aforementioned are extracted from satellite
images and used as clues to locate
regions where thunderstorms tend to happen in the near future.
Machine learning is involved in the classification of
candidate storm regions by inspecting historical thunderstorm data.
The regions, as extracted from synoptical scale features,
are not necessarily associated one-to-one with storms,
i.e., a region may contain several storms and each storm needs
to be more precisely located by other techniques. The algorithm
in fact simulates meteorologists' cognition process
when reviewing satellite images as a part of the weather prediction
process. It could be a helpful tool in automating
and refining the current weather prediction. It can also be used
to provide assistance to meteorologists so that they can be more efficient 
and less exhausting.

\subsection{Outline}
The rest of the paper is organized as follows. Section~\ref{sec:feature_extract} introduces
the approach to extract storm signatures and construct storm features. Section~\ref{sec:machine_learning}
reports the approach and results of the machine learning module that accepts or rejects the
extracted storm features. In particular, quantitative benchmarks of the classifier and qualitative
case studies that applies the whole workflow are given to demonstrate the effectiveness of the
proposed algorithm. Finally, we present the future work and conclude the paper in Section~\ref{sec:conclusion}.


\section{Storm Feature Extraction}\label{sec:feature_extract}
Our algorithm extracts synoptic-scale visual thunderstorm features
from satellite images. The cloud motion observed
from image sequences is analyzed in depth.
In order to simulate human cognition and let computers
perceive comma shapes introduced earlier,
the visual signatures are further abstracted into
basic cloud motion components: translation and rotation.
Several properties and measurements related to these two
components are extracted from cloud
motions and analyzed in the following steps.

\begin{figure}
\includegraphics[width=\textwidth]{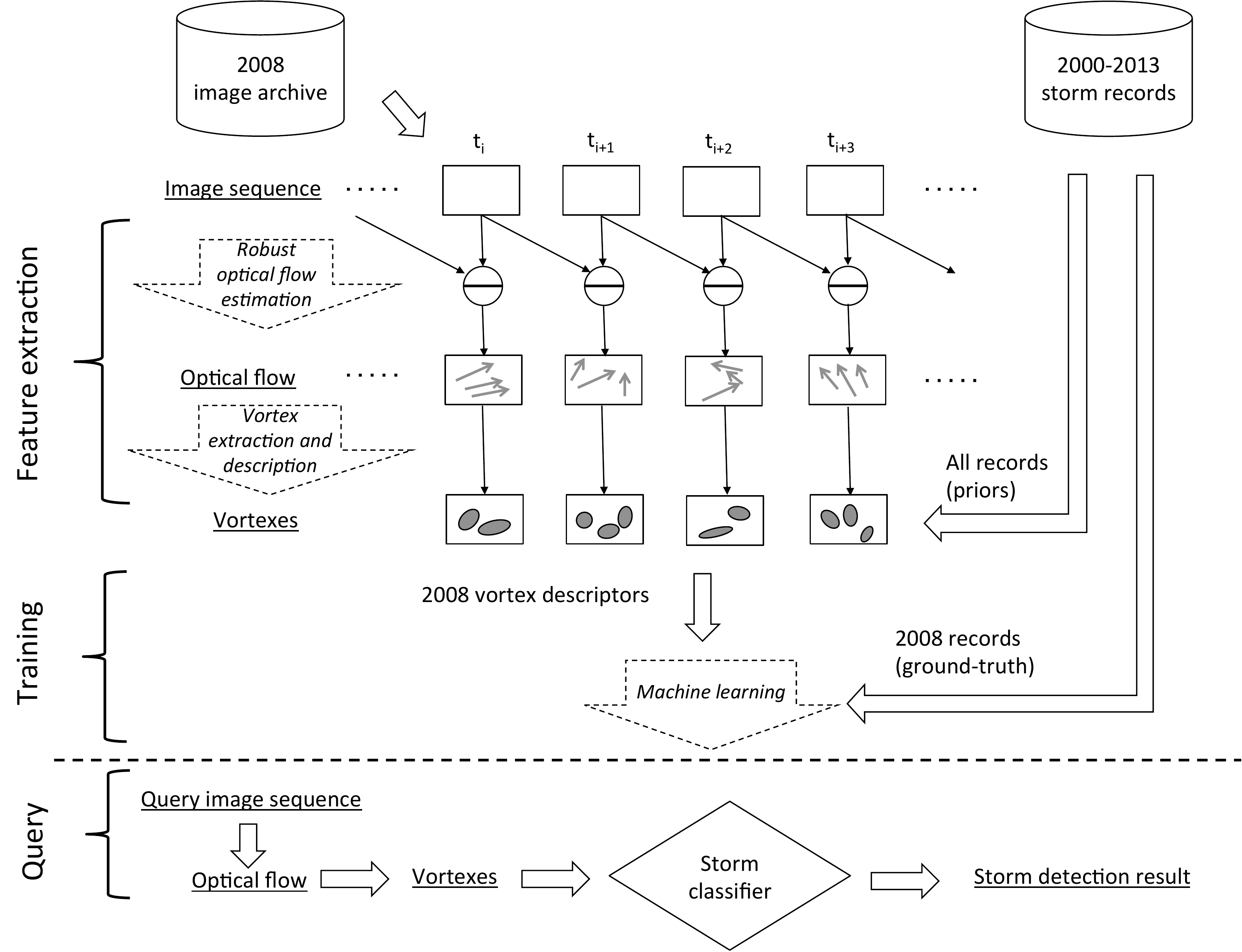}
\caption{Workflow of the storm detection system. Optical flows
between adjacent image frames are estimated and vortexes are
extracted base on the flows. Vortex descriptors are then
constructed using both the visual information and historical
storm records. Machine learning is performed on the vortex
descriptors for the storm detection.
}\label{fig:framework}
\end{figure}

Fig.~\ref{fig:framework} illustrates the workflow of the whole system.
We employ the {\it optical flow}
between every two adjacent satellite images
to capture the cloud motion and discover vortex
areas that could potentially trigger storms.
Using the historical storm records, vortex features are constructed
and a storm classifier is trained. Given an arbitrary query image
sequence, vortexes are extracted in the same way and then categorized
by the classifier.

In particular, for the feature extraction operations, which will be
applied both to the training data and the query data, two steps are
carried out.
The system first estimates a dense optical flow field
describing the cloud motion between adjacent image frames.
Second, local vortexes are identified with the optical flow
and vortex descriptors are constructed. The vortex descriptors
combine information from both the visual
features and the historical storm records.
The following subsections describe details of these steps.

\subsection{Robust Optical Flow Estimation}\label{sec:optical_flow}
Optical flow is a basic technique for motion estimation
in image sequences. Given two images $I_t(x,y)$ and
$I_{t+1}(x,y)$, the optical flow $\vec{F}_t(x,y)=\{U_t(x,y),V_t(x,y)\}$
defines a mapping for each pixel
$g:(x,y) \mapsto (x+U_t(x,y), y+V_t(x,y))$, so that
$I_t(g(x,y))\approx I_{t+1}(x,y)$. The vector field
$\vec{F}_t(x,y)$ can be therefore regarded as
the pixel-wise motions from image $I_t(x,y)$ to
$I_{t+1}(x,y)$. Several approaches for optical
flow estimation have been proposed based on different
optimization criteria~\cite{horn1981determining,lucas1981iterative,perez2013tv}. In this work we adopt
the pyramid Lucas-Kanade algorithm~\cite{bouguet2001pyramidal} to estimate a
dense optical flow field between every two neighboring
image frames.

\begin{figure*}
\centering
\subfigure[Noisy optical flow estimation]{
    \includegraphics[width=0.45\textwidth]{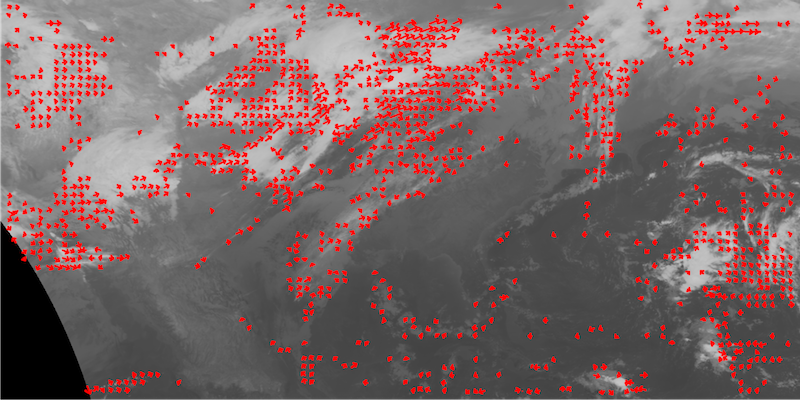}
    \label{fig:process_1}
}
\subfigure[Smoothed optical flow]{
    \includegraphics[width=0.45\textwidth]{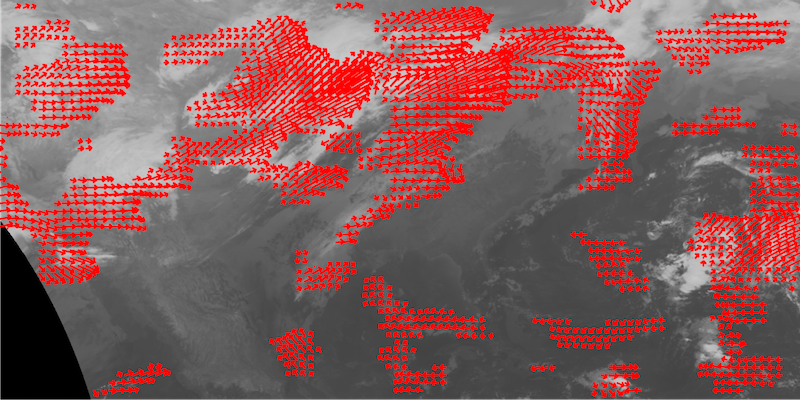}
    \label{fig:process_2}
}\\
\subfigure[Divergence-free component of the smoothed flow]{
    \includegraphics[width=0.45\textwidth]{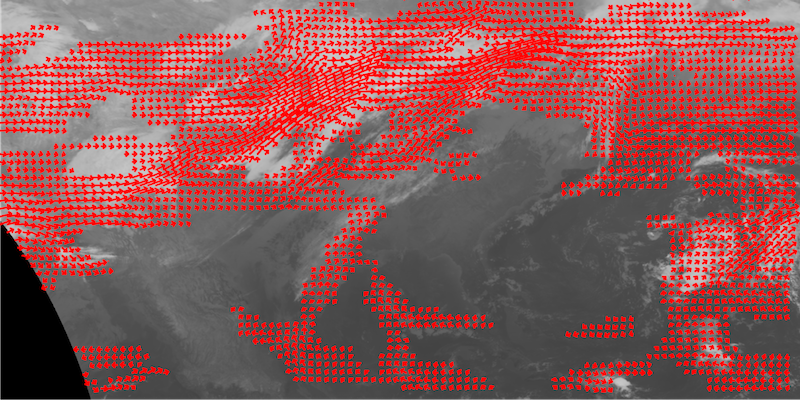}
    \label{fig:process_3}
}
\subfigure[Vorticity-free component of the smoothed flow]{
    \includegraphics[width=0.45\textwidth]{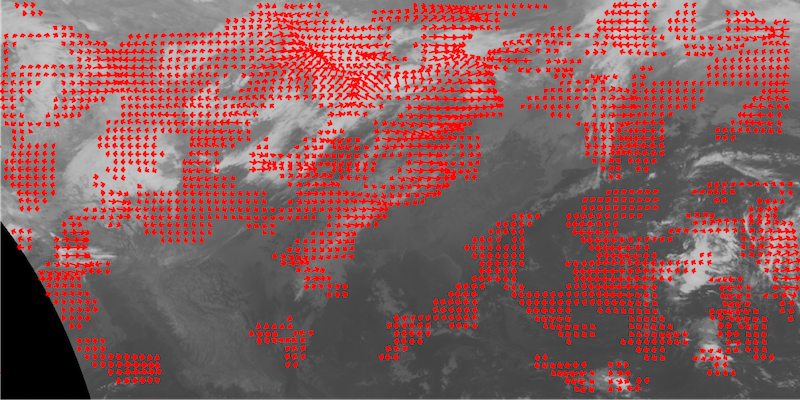}
    \label{fig:process_4}
}\\
\subfigure[Magnitude of vorticity]{
    \includegraphics[width=0.45\textwidth]{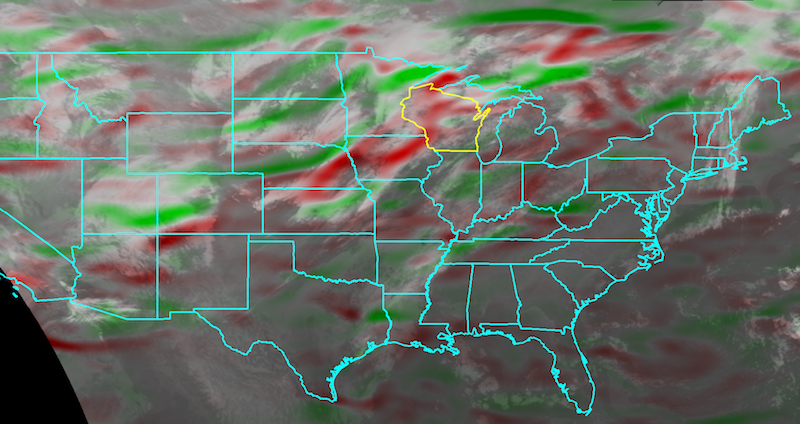}
    \label{fig:process_5}
}
\subfigure[Vortex cores and their expansions]{
    \includegraphics[width=0.45\textwidth]{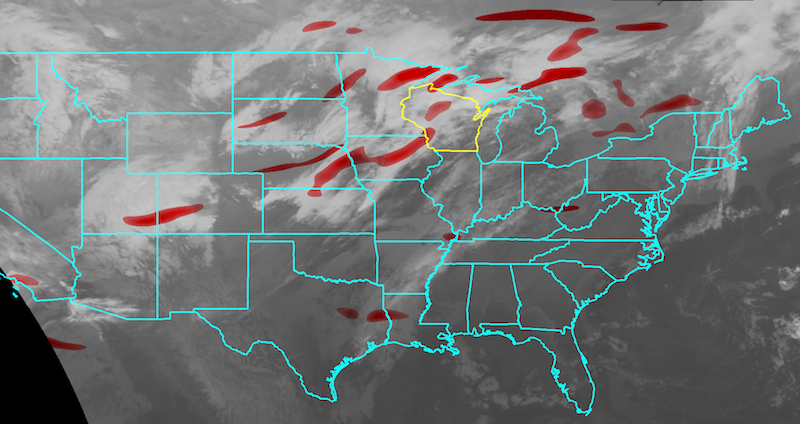}
    \label{fig:process_6}
}
\caption{Sample results of the optical flow analysis.
GOES-M satellite images covering the continental US area
are analyzed (to avoid showing too many 
missing pixels on the southwest corner, a
small stripe of US west coast is not shown on the figures, though
these pixels are included in our analysis).
Between two adjacent
frames (with 30 minutes' interval) in a satellite image sequence, the optical
flow from the former frame to the latter
one is estimated and processed.
The results are plotted on the second frame.
Only large enough flow vectors are drawn on figure (a-d).
(a) The optical flow estimated
by the Lucas-Kanade algorithm. (b) The smoothed optical flow by applying
an iterative update based on the Navier-Stokes equation. (c) The
divergence-free component of the smoothed flow field. (d) The vorticity-free
component of the smoothed flow field. (e) Visualization of the vorticity.
Vorticity vectors of the 2D flow field are perpendicular to the image plane.
Pixels with vorticity vectors toward the viewer (counter-clockwise rotations)
are tinted in green color; and the pixels with vorticity away from the
viewer (clockwise rotations) are tinted in red color. The saturation
of color means the magnitude of the corresponding vorticity vector.
(f) Vortex regions detected by the $Q$-criterion. For visualization
purpose (to avoid displaying too a large missing corner at
the lower-left, not all areas in CONUS are shown on the above images.
}
\end{figure*}

The GOES-M satellite is in a geostationary orbit and thus remains over
the same point on the earth at all times. As a result,
the only objects moving along a satellite image sequence are 
the clouds and the optical flow between two images indicates
the cloud motion.
However, the non-rigid and dynamic nature of clouds
makes the optical flow estimation noisy and inaccurate.
Fig.~\ref{fig:process_1} shows the optical flow
estimation result of a satellite
image in respect to its previous frame in the sequence.
Though the result correctly reflects the general cloud
motion, flows in local areas are usually noisy and do not
precisely describe the local motions.
As to be introduced later, the optical flow properties
adopted in this work involve the gradient operation of the
flow field. Thus it is important to get reliable and smooth
optical flow estimation.
To achieve this goal, we
both pre-process the images and post-process the estimation
results. Before calculating the optical flow between
two images, we first enhance both of them using the histogram
equalization technique\footnote{
Each channel is enhanced separately. On a given channel,
the same equalization function (estimated from the first frame
of the sequence) is applied to both images
so that they are enhanced by the same mapping.}~\cite{acharya2005image}.
As a result, more fine details on the images are enhanced
for better optical flow estimation.

After the initial optical flow estimation,
we smooth the optical flow by applying an iterative update operation based
on the Navier-Stokes equation~\cite{temam1984navier} for modeling fluid motions. Given
a flow field $\vec{F}(x,y)$, the equation describes the
evolving of the flow over time $t$:
\begin{equation}\label{equ:ns}
\frac{\partial\vec{F}}{\partial t}=(-\vec{F}\cdot\nabla)\vec{F}+\nu\nabla^2\vec{F}+\vec{f}\;,
\end{equation} 
where $\nabla=(\frac{\partial}{\partial x},\frac{\partial}{\partial y})$ is the 
gradient operator, $\nu$ is the viscosity of the fluid, and $\vec{f}(x,y)$ is the 
external force applied at the location $(x,y)$.

The three terms in Eq.~(\ref{equ:ns}) correspond to the advection, diffusion, and
outer force of the flow field respectively. In the method
introduced in~\cite{stam1999stable}, an initial flow field is updated
to a stable status by iteratively applying these three transformations one
by one. Compared with a regular low-pass filtering to the flow field,
this approach takes into account the physical model of fluid dynamics,
and therefore better approximates the real movement of a flow field.
We adopt a similar strategy to smooth
the optical flow in our case.
The initial optical
flow field $\vec{F_t}(x,y)$ is iteratively updated.
Within each iteration the time is incremented by a
small interval $\Delta t$, and there are three steps to get
$\vec{F}_{t+\Delta t}(x,y)$ from $\vec{F}_t(x,y)$:
\begin{enumerate}
\item add force: $\vec{F}_{t+\Delta t}^{(1)}=\vec{F}_t+\vec{f}\Delta t$\;;
\item advect: $\vec{F}_{t+\Delta t}^{(2)}=adv(\vec{F}_{t+\Delta t}^{(1)}, -\Delta t)$\;;
\item diffuse: $\vec{F}_{t+\Delta t}=FFT^{-1}(FFT(\vec{F}_{t+\Delta t}^{(2)})e^{-\nu k^2\Delta t})$\;.
\end{enumerate}

The first step is simply a linear increment of the flow vectors based on the
external force. In the second step, the advection operator $adv(\cdot)$ uses
the flow $\vec{F}_{t+\Delta t}^1(x,y)$ at each pixel $(x,y)$ to predict its
location $(x_{t-\Delta t}, y_{t-\Delta t})$ $\Delta t$ time ago, and updates
$\vec{F}_{t+\Delta t}^{(2)}(x,y)$ by $\vec{F}_{t+\Delta t}^{(1)}(x_{t-\Delta t}, y_{t-\Delta t})$.
In the last step, the diffusion operation is a low-pass
filter applied in the frequency domain,
where $k$ is the distance from a point to the origin, and the bandwidth of
the filter is determined by the pre-defined fluid viscosity $\nu$ and $\Delta t$ (to be specified later).
We do not enforce the mass conservation like the approach in~\cite{stam1999stable}
because the two-dimensional flow field corresponding to the cloud movement
is not necessarily divergence free.
In fact, we use the divergence as a feature for storm detection
in the following procedures.

After several iterations of the update, the flow field converges to a
stable status and becomes smoother. The iteration number is typically not large,
and we find that the final result is not very sensitive to the parameters
$\nu$ and $\Delta t$. In our system we set the iteration number to 5, the fluid
viscosity $\nu$ to 0.001, and time interval $\Delta t$ to 1. The noisy
optical flow estimation from the previous stage is treated as the external
force field $\vec{f}$, and initially $\vec{F}_t = 0$.
Fig.~\ref{fig:process_2} shows the smoothed flow field
(i.e., $\vec{F}_{t+5\Delta t}$) from the noisy
estimation (Fig.~\ref{fig:process_1}).
Clearly the major motion information is kept
and the flow field is smooth for further analysis.

\subsection{Flow Field Vortex Extraction}
As introduced in Section~\ref{sec:intro}, the rotating and diverging of
local cloud patches are two key signatures for storm detection. In the flow
field, these two kinds of evidence are embodied by the {\it vorticity} and
{\it divergence}. The vorticity of a vector field $\vec{F}(x,y)=\{U(x,y),V(x,y)\}$ is defined as:
\begin{equation}
\begin{aligned}
\vec{\omega}(x,y)&=\nabla \times \vec{F}(x,y)\\
&=\left(\frac{\partial}{\partial x},\frac{\partial}{\partial y},\frac{\partial}{\partial z}\right)\times(U(x,y),V(x,y),0)\\
&=\left(\frac{\partial V(x,y)}{\partial x}-\frac{\partial U(x,y)}{\partial y}\right)\vec{z}\;;
\end{aligned}
\end{equation}
and the divergence is defined as:
\begin{equation}
\begin{aligned}
\text{div}\vec{F}(x,y) &= \nabla \cdot \vec{F}(x,y)\\
&=\left(\frac{\partial}{\partial x},\frac{\partial}{\partial y}\right)\cdot(U(x,y),V(x,y))\\
&=\frac{\partial U(x,y)}{\partial x}+\frac{\partial V(x,y)}{\partial y}\;.
\end{aligned}
\end{equation}

It has been proven that for a rigid body, the magnitude of vorticity at any
point is twice the angular velocity of its self
rotation~\cite{green1995fluid} (the direction of the vorticity vector indicates
the rotation's direction). In our case, even though the cloud is non-rigid,
we can regard the vorticity at a certain point as a description of the local rotation.

To better reveal the rotation and divergence, we apply the 
Helmholtz-Hodge Decomposition~\cite{arfken1999mathematical} to decompose the flow field to
a solenoidal (divergence-free) component and a irrotational (vorticity-free)
component. Fig.~\ref{fig:process_3} and Fig.~\ref{fig:process_4} visualize
these two components respectively. In both figures,
areas with densely overlapped vectors plotted are
the places with high vorticity or divergence.

The divergence-free component of the flow field is useful for
detecting vortexes. On this component, we inspect the
local deformation tensor
$$
\nabla \vec{F} = \left[
\begin{matrix}
  {\partial U}/{\partial x} & {\partial V}/{\partial x} \\
  {\partial U}/{\partial y} & {\partial V}/{\partial y}
 \end{matrix}
\right]\;,
$$
which can be decomposed to the symmetric (strain) part
$
S=\frac{1}{2}({\nabla \vec{F} + \nabla \vec{F}^T})
$
and the asymmetric (rotation) part
$
\Omega=\frac{1}{2}({\nabla \vec{F} - \nabla \vec{F}^T})
$.
The $Q$-criterion introduced by~\cite{hunt1988eddies}
takes the difference of their norms:
\begin{equation}\label{equ:Q-cri}
Q=\frac{1}{2}(\|\Omega\|^2-\|S\|^2)=\frac{1}{4}\|\vec{\omega}\|^2-\frac{1}{2}\|S\|^2\;,
\end{equation}
where $\|\cdot\|$ is the Frobenius matrix norm.
The $Q$-criterion measures the dominance of the vorticity.
When $Q>0$, i.e., the vorticity component dominates the local flow, the
corresponding location is regarded to be in a vortex region.
Fig.~\ref{fig:process_5} and Fig.~\ref{fig:process_6} visualize
the vorticity and $Q$-criterion of the flow field in Fig.~\ref{fig:process_3}.
In Fig.~\ref{fig:process_5}, pixels are tinted by the corresponding
magnitude of vorticity, and different
colors mean different rotating directions.
In Fig.~\ref{fig:process_6}, the vortex regions are highlighted in red.
Clearly only pixels with a dominant vorticity component are selected 
as vortexes by the $Q$-criterion.
It is apparent that these vortexes are more prone to be located
inside the storm area (highlighted by yellow boundaries)
than the removed high-vorticity pixels.
It is also observed that the vortexes are typically in narrow
bending comma shapes as described in Section~\ref{sec:pattern}.
Therefore, they are properly to be regarded as the potential storm elements.

\subsection{Vortex Descriptor}\label{sec:fea}

Not all the vortex regions in Fig.~\ref{fig:process_6} are related to
storms\footnote{The types of
vortexes outside of the CONUS are unknown because of lack of records.}.
As a result we built a descriptor for each extracted vortex and
apply it to a machine learning module (to be introduced later).
We introduce the vortex descriptor in this subsection.

Denoting a certain vortex region in a satellite image as $\Phi$ and the area
of $\Phi$ as $\Pi(\Phi)$, the following
visual clues, both static ones from a single image and dynamic ones from
the optical flow with respect to the previous image frame, are considered
in our approach.
\begin{enumerate}
\item Mean channel 3 brightness:
$$
w_1(\Phi) = \frac{\sum_{(x,y)\in\Phi}I^{(3)}(x,y)}{\Pi(\Phi)}\;,
$$
where $I^{(3)}(x,y)$ is the brightness of pixel $(x,y)$ in the channel 3 image.
\item Mean channel 4 brightness:
$$
w_2(\Phi) = \frac{\sum_{(x,y)\in\Phi}I^{(4)}(x,y)}{\Pi(\Phi)}\;,
$$
where $I^{(4)}(x,y)$ is the brightness of pixel $(x,y)$ in the channel 3 image.
\item Mean optical flow intensity:
$$
w_3(\Phi) = \frac{\sum_{(x,y)\in\Phi}\|\vec{F}(x,y)\|}{\Pi(\Phi)}\;,
$$
where $\vec{F}(x,y)=(U(x,y), V(x,y))$ is the optical flow at pixel $(x,y)$ computed between the previous and current 
image frames.
\item Mean optical flow direction:
$$
w_4(\Phi) = \frac{\sum_{(x,y)\in\Phi}\|\theta(x,y)\|}{\Pi(\Phi)}\;,
$$
where $\theta(x,y)=\tan^{-1}\left(\frac{V(x,y)}{U(x,y)}\right)$ is optical flow $\vec{F}(x,y)$'s direction.
\item Mean vorticity on the solenoidal optical flow component:
$$
w_5(\Phi) = \frac{\sum_{(x,y)\in\Phi}\omega(x,y)}{\Pi(\Phi)}\;,
$$
where $\omega(x,y)$ is the vorticity value of the solenoidal component of
optical flow $\vec{F}(x,y)$ (sign of the value
indicates the vorticity's direction).
\item Mean divergence on the irrotational optical flow component:
$$
w_6(\Phi) = \frac{\sum_{(x,y)\in\Phi}\text{div}\vec{F}(x,y)}{\Pi(\Phi)}\;,
$$
where $\vec{F}(x,y)$ is the divergence of the irrotational component of
optical flow $\vec{F}(x,y)$.
\item Maximal $Q$-value of the vortex:
$$
w_7(\Phi) = \max_{(x,y)\in\Phi}{Q(x,y)}\;,
$$
where $Q(x,y)$ is defined in Eq.~(\ref{equ:Q-cri}).
\end{enumerate}

\begin{figure*}
\centering
\includegraphics[width=0.3\linewidth]{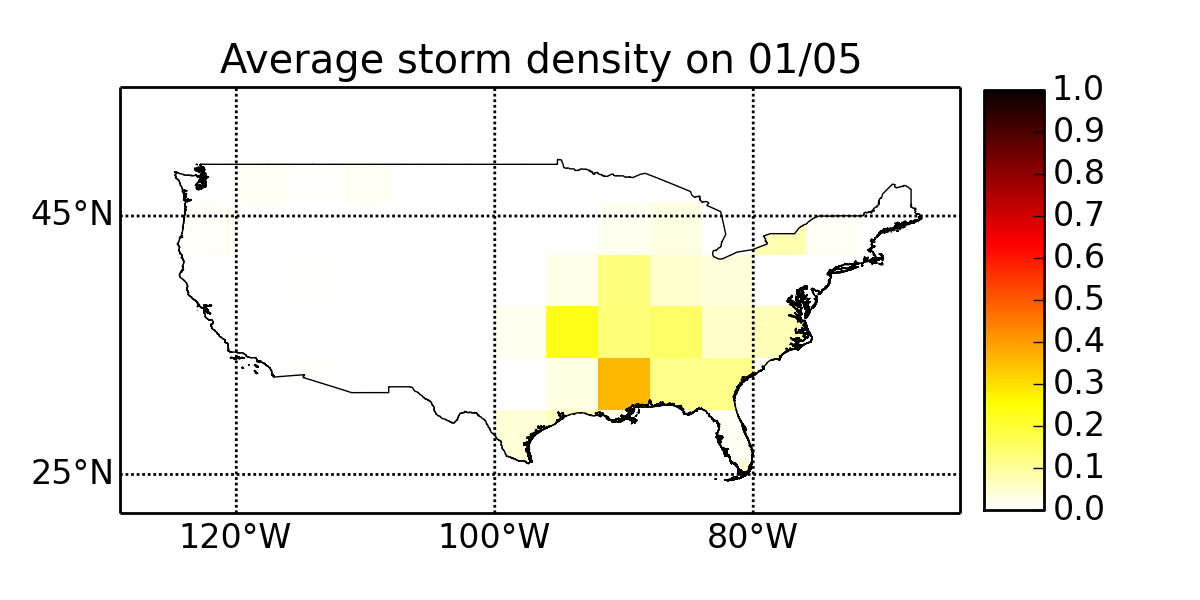}
\includegraphics[width=0.3\linewidth]{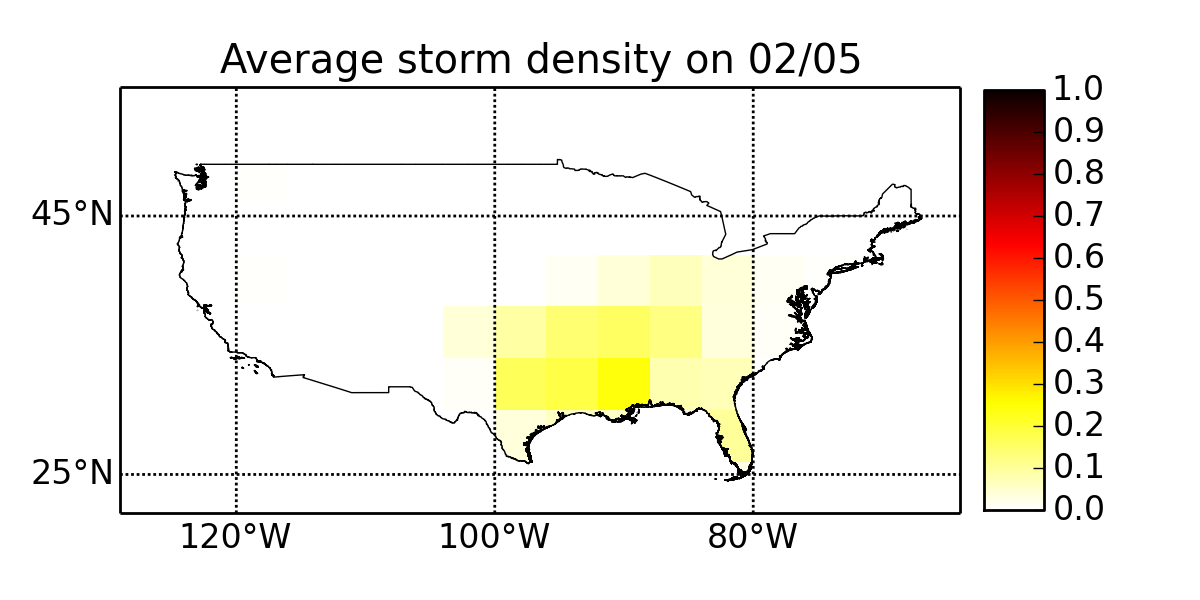}
\includegraphics[width=0.3\linewidth]{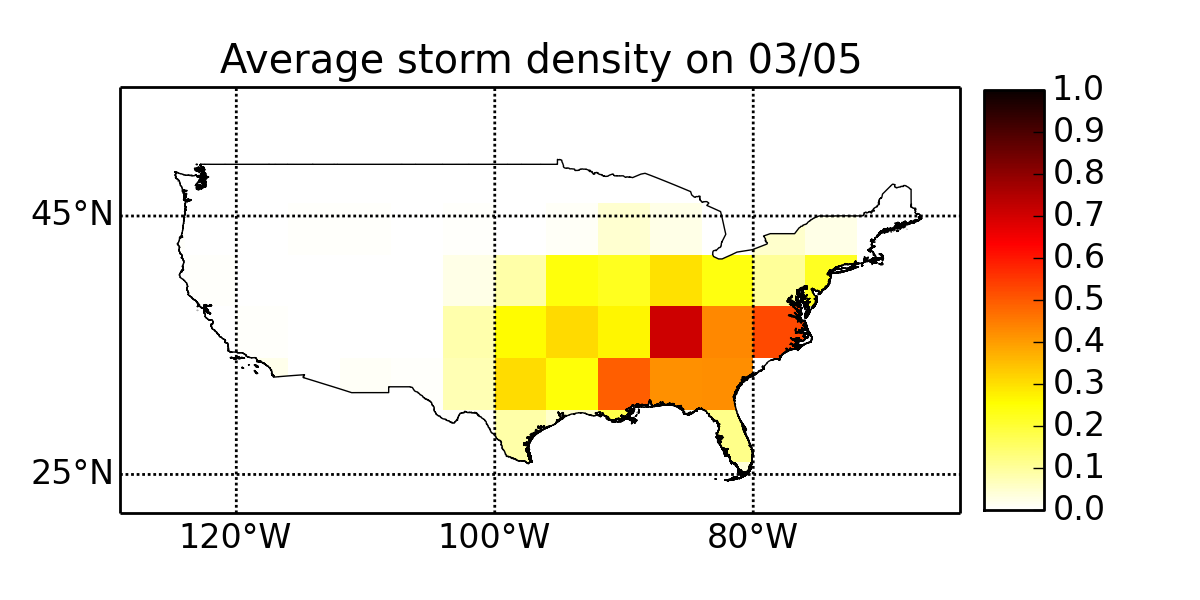}\\
\includegraphics[width=0.3\linewidth]{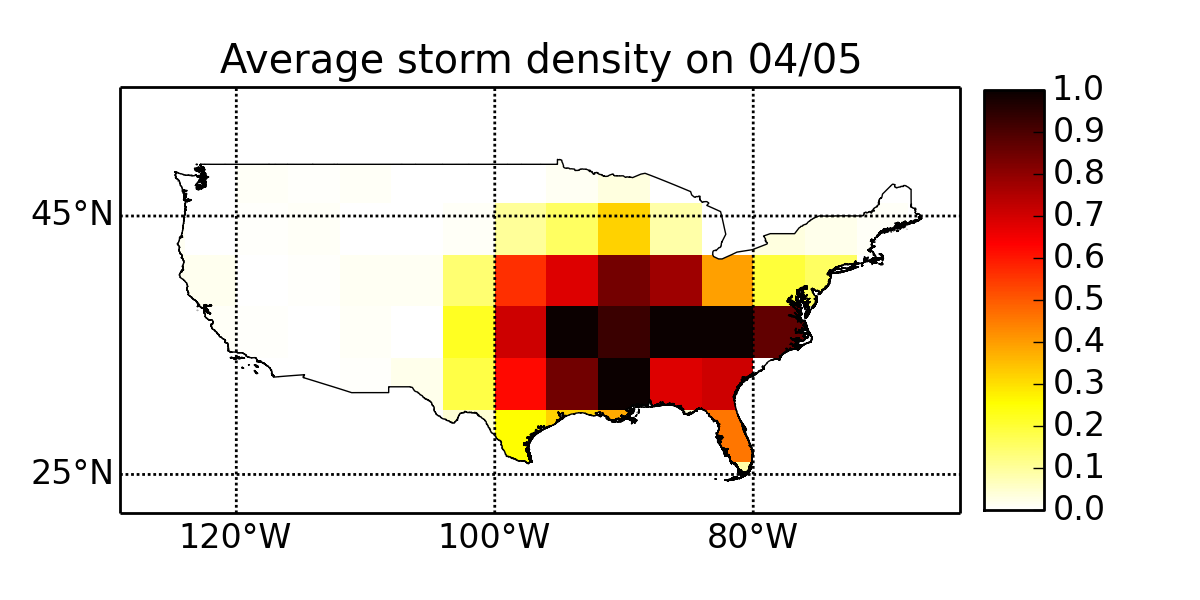}
\includegraphics[width=0.3\linewidth]{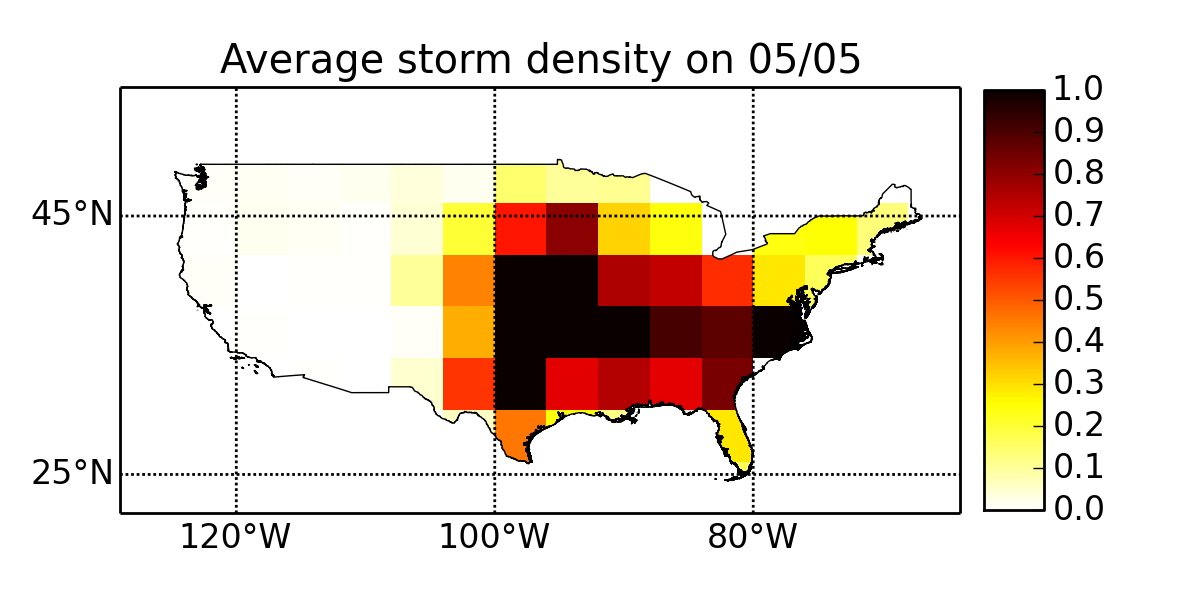}
\includegraphics[width=0.3\linewidth]{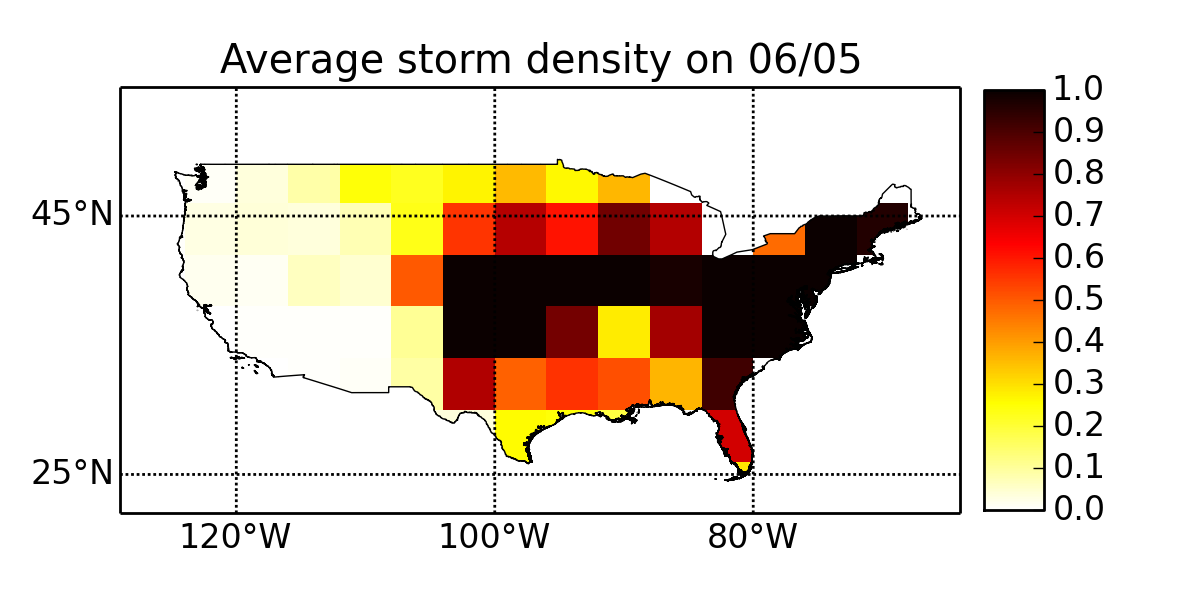}\\
\includegraphics[width=0.3\linewidth]{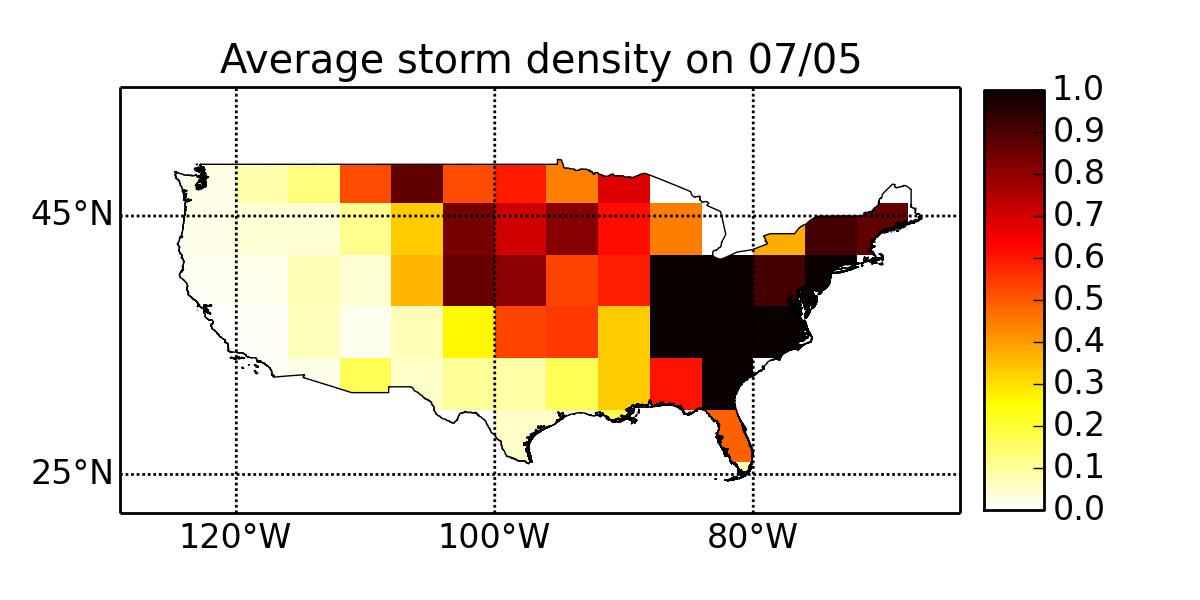}
\includegraphics[width=0.3\linewidth]{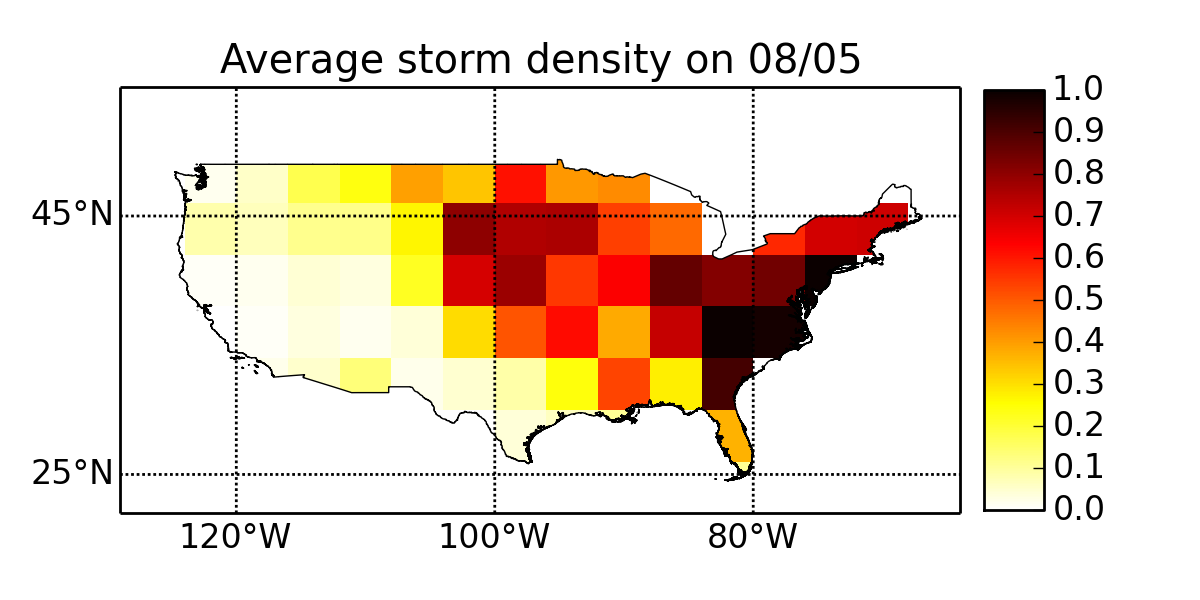}
\includegraphics[width=0.3\linewidth]{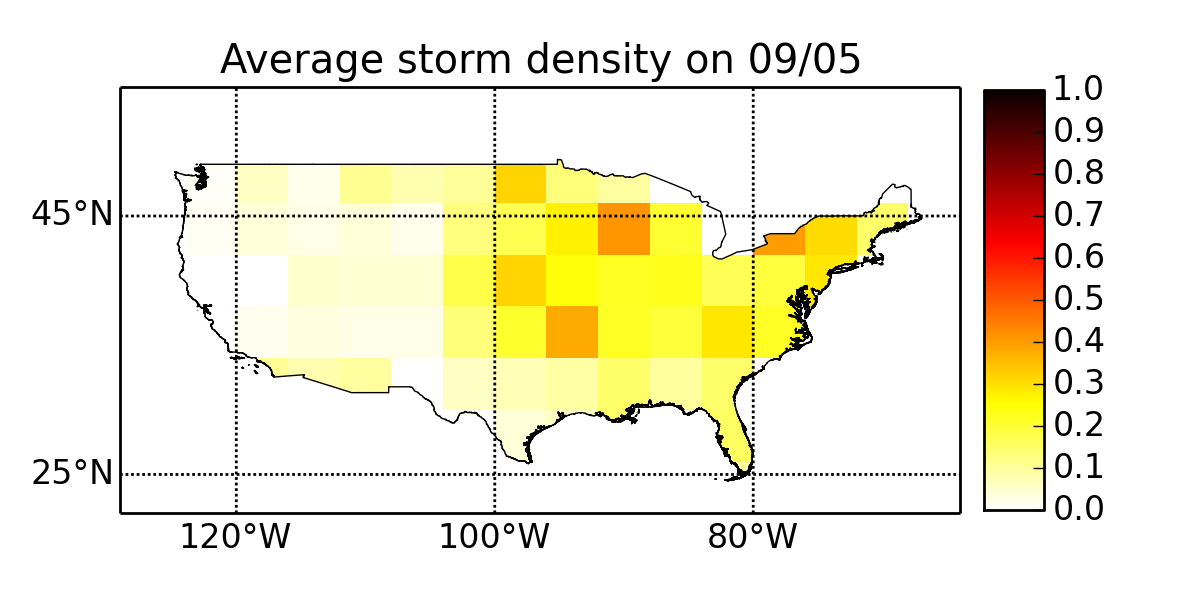}\\
\includegraphics[width=0.3\linewidth]{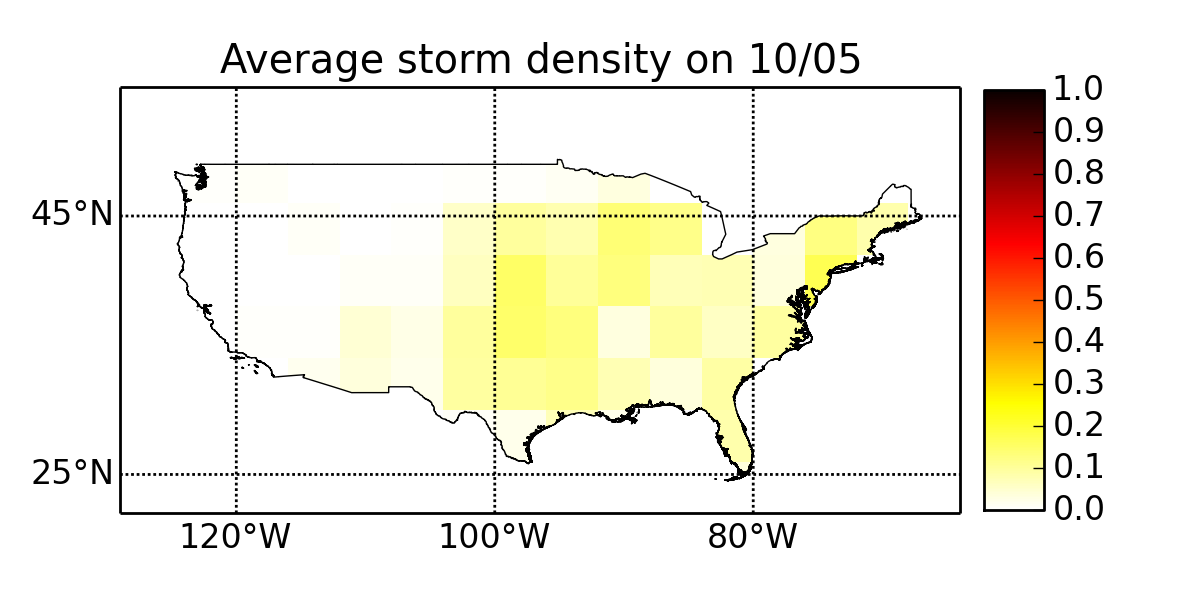}
\includegraphics[width=0.3\linewidth]{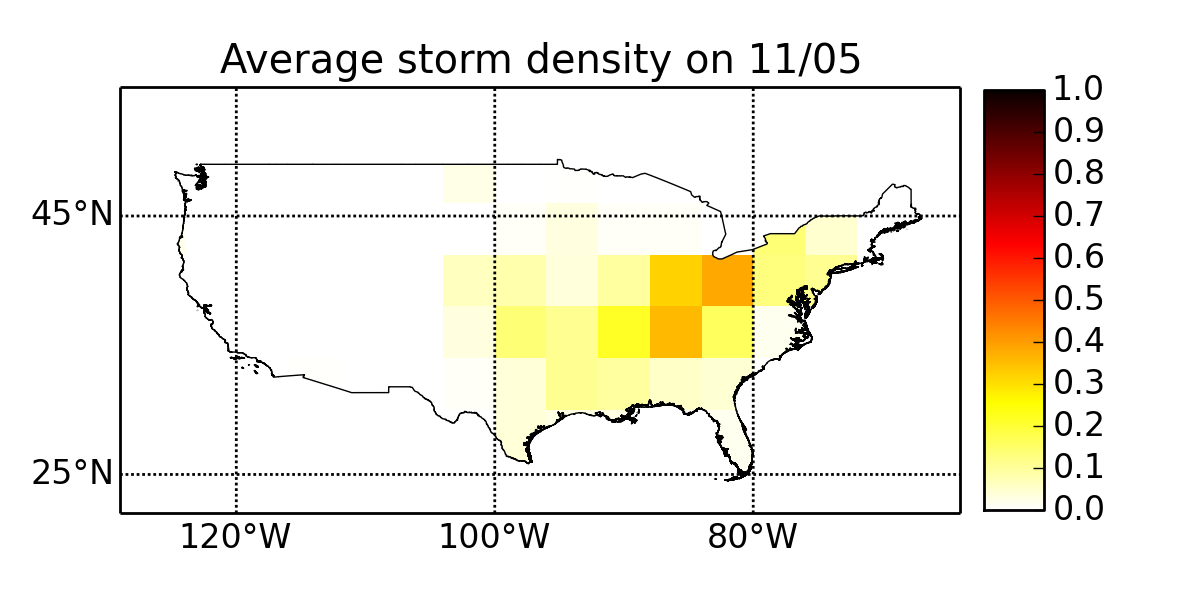}
\includegraphics[width=0.3\linewidth]{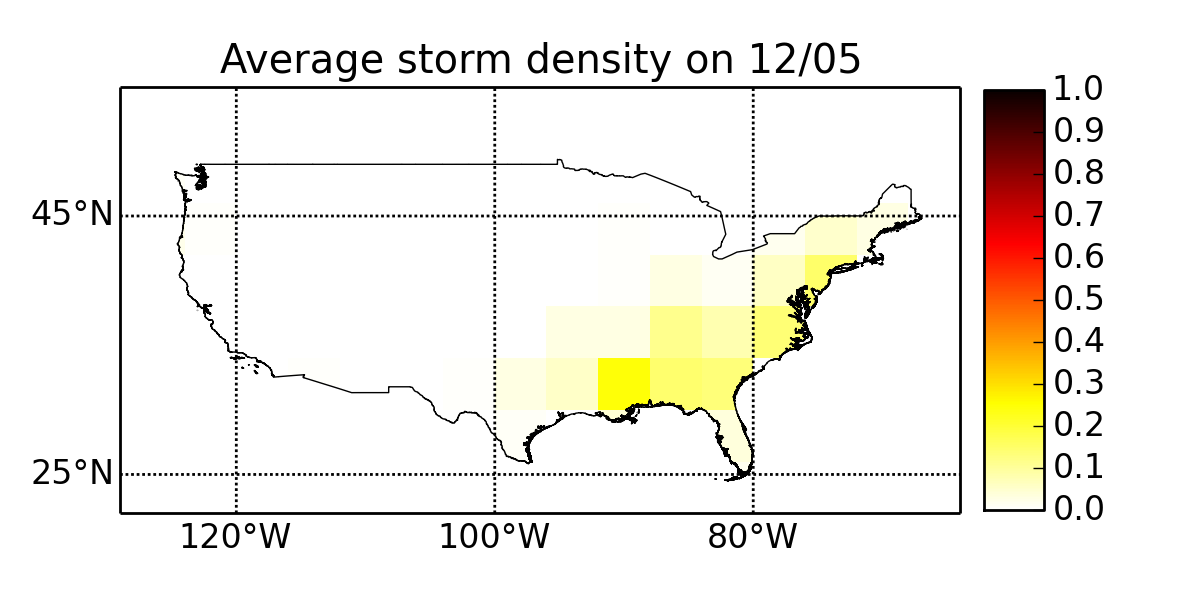}

\caption{
Average storm densities (number of storms
per 30,000~$\text{km}^2$) for the fifth day of each month in US
continent. Darker color means higher storm density.
The statistics are based on the historical records
from 2000 to 2013. Around each given date (+/-5 days), we count the total
number of storms reported in each state across the past 14 years.
The numbers are then divided by the total number of days and the areas of
corresponding states to calculate the average densities.}\label{fig:dense}
\end{figure*}

We also considered the spatial and temporal distributions of the storms that
occurred in the CONUS from 2000 to 2013 using all the historical
storm records through these years\footnote{We noticed that
NWS revised the criteria for hail reporting from 3/4 to 1 inch in
diameter in 2010 so less hails were reported after that.
However, the change has little effect on the storms' relative spatial distribution used in the machine learning system.
}.
This is shown in Fig.~\ref{fig:dense}. We divided the CONUS area into
$4^{\circ} \times 4^{\circ}$ grids. Inside each grid, we count the occurrence
of storms around each given date (e.g., July 4th) in the history.
Storms that occurred from 5 days before to 5 days after the queried
date in each year are counted (e.g., for July 4, storms
from June 29 to July 9 in every year, a total of 154 days,
are included\footnote{We count storms around February 28th in non-leap years
for storm statistics on February 29th.}). Denoting the total
storm count in grid $(i,j)$ (the grid counted $i$-th from the left and
$j$-th from the top) for date $d$ is $N_d(i,j)$
, the average storm density in the grid is
$\rho_d(i,j) = \tfrac{N_d(i,j)}{154}$.

To describe the storm prior for vortex $\Phi$
(extracted from date $d$), we took the average storm densities of
the pixels it covers:
$$
w_8(\Phi) = \frac{\sum_{(x,y)\in\Phi}\phi_d(T(x,y))}{\Pi(\Phi)}\;,
$$
where $T:(x,y) \mapsto (i, j)$ is a mapping from the pixel $(x,y)$ to
the grid index $(i,j)$ it belongs to.

With all the features introduced above, for vortex $\Phi$ we constructed
a vortex descriptor $X(\Phi)=(w_1, w_2, w_3, w_4, w_5, w_6, w_7, w_8)$.
Descriptors of training data are fed into a machine learning algorithm,
and eventually we obtain a classifier $C(\cdot)$.
For an arbitrary vortex $\Phi$, $Y(\Phi) = C(X(\Phi))$
is the predicted status calculated by the classifier. We introduce
the machine learning approach in the next section.

\section{Storm System Classification and Detection}\label{sec:machine_learning}
The vortex regions with large $Q$ values correspond
to regions with strong cloud rotations and our approach extracts them as vortexes.
However, they are not necessarily related to the storms because
the cloud rotation is only one aspect of a weather system.
In order to create reliable storm detection,
we embedded these vortexes and 
their descriptors into a machine learning framework.
Using the descriptors and ground-truth labels
of the extracted candidate vortex regions,
a classifier was trained to distinguish storm-related
vortexes from other vortexes.

\subsection{Training Vortexes and Ground-truth Labels}~\label{sec:gnd_truth}

We extracted vortexes from 2008 GOES-M satellite images and used the storm
reports in the same year to assign a ground-truth label for each vortex
in the CONUS. A classifier is trained to associate the vortex features
with the storm labels. Image sequences in the first ten days of each
month are used for training and those from the 18th to the 22nd days
of each month are used for testing (so that the weather systems
in the training and testing data are relatively far apart and independent). In order to get enough difference yet not to lose
too much detail, we choose the
sampling interval between every two images in a sequence to be
30 minutes.

We used the historical storms to assign ground-truth labels to detected vortexes.
We denote the historical storm database as $\mathscr{D}$ and each storm
entry as $(u_i, v_i, t_i)$ where $(u_i,v_i)$ is the location
(latitude and longitude values in degrees) and $t_i$ 
is the starting time of the storm. 
Assuming a vortex is detected at location $(u, v)$\footnote{$(u, v)$ is
the geometric center of the vortex.} and time $t$,
if at least one record $(u_i, v_i, t_i)\in\mathscr{D}$ within $3^{\circ}$ of
latitude and longitude
exists between the moments $t-0.5hr$ and $t+2hr$
(i.e. $\|u_i-u\| < 3^{\circ} \text{ and } \|v_i-v\| < 3^{\circ} \text{ and }t-0.5hr\le t_i \le t+2hr$),
the vortex is assigned with a positive (storm-related) label;
otherwise it is assigned with a negative (no-storm) label.
We only assigned labels for vortexes inside the CONUS.
For vortexes outside of the CONUS, we had no historical data
indicating whether or not they are storm-related, so we did not
use them for training and testing.

On the 120 selected days (10 for each month), a total of 8,527
storm-related vortexes were discovered by the above method. We
randomly selected the same number of vortexes in the CONUS
that were not storm-related from these days.
A binary classifier is then trained on the 17,054 training vortex samples.

\subsection{Random Forest Classifier}\label{sec:training}
We trained a random forest classifier~\cite{breiman2001random}
using the training data to distinguish storm-related vortexes
from vortexes not affiliated with a storm.
The features for each vortex are
described in Section~\ref{sec:fea} and the ground-truth is defined in
Section~\ref{sec:gnd_truth}.

Using the training data, the random forest learns rules
to determine whether a vortex of certain descriptor causes storms.
The random forest is essentially an ensemble learning
algorithm that makes predictions by combining the results of
multiple individual decision trees trained from random subsets
of the training data.
We chose this approach because it has been found to outperform a single
classifier (e.g., SVM).  In fact, this strategy resembles the {\it
  ensemble forecasting}~\cite{molteni1996ecmwf} approach in
numerical weather prediction, where multiple numerical results with
different initial conditions are combined to make a more reliable
weather forecast.  For both numerical and statistical weather
forecasting approaches, the ensemble approaches help to improve
the prediction qualities.

\subsection{Testing Vortex Samples and Testing Methods}
To develop a benchmark for the performance of the vortex classifier,
we used image sequences from the 18th to the 22nd days
of each month in 2008 as the testing data.
On each selected day,
the image sequence (with a 30-minute interval)
from 10:00~GMT to 18:00~GMT is used to extract dynamic
cloud vortexes.
The test dataset contains a total of 48,698 vortexes,
among which 781 are labeled as positive, and the rest 47,917
samples are negative.
The descriptors of these vortexes are classified
by the classifier. We then evaluate the classification results
in two ways.

Firstly, we assigned the testing vortexes with ground-truth
labels in the same way as how the training ground-truth labels are assigned.
The predicted storm status for each test vortex
is compared with the ground-truth. In this way the classification
accuracy for single vortexes are evaluated. Table~\ref{tab:cls} presents
the classification performance of both the training data (by cross validation)
and the testing data. The classifier shows
consistent performances on both the training set and the test set.
To demonstrate the effect of including geographic storm priors in
the classification, we also trained and tested the random forest classifiers
only on the visual features and the storm density separately.
Clearly none of the visual features and
the historical storm density standalone performs well alone.
Combining visual features with the storm density
significantly enhances the classification performance.

\begin{table}
\renewcommand{\arraystretch}{1.3}
\caption{Classification Performance and Contributions of Visual Features and Historical Priors}\label{tab:cls}
\begin{center}
\begin{tabular}{|c|c|p{0.15\columnwidth}|p{0.15\columnwidth}|p{0.15\columnwidth}|}
\hline
\multicolumn{2}{|c|}{ }& All features & Visual only & Prior only\\\hline
\multirow{3}{*}{\parbox[t]{0.2\columnwidth}{Training set \\(cross validation)}} 
& Overall& 83.3\%& 68.8\%& 76.1\%\\\cline{2-5}
& Sensitivity &89.6\%& 67.4\%& 74.8\% \\\cline{2-5}
& Specificity & 76.9\%& 70.3\%&77.3\%\\\hline
\multirow{3}{*}{\parbox[t]{0.2\columnwidth}{Testing set}}
& Overall& 80.3\% & 67.4\%& 77.3\%\\\cline{2-5}
& Sensitivity & 82.1\%& 47.2\%& 77.4\%\\\cline{2-5}
& Specificity & 80.2\%& 67.7\%& 71.2\%\\\hline
\multicolumn{5}{c}{ }
\end{tabular}
\end{center}
{
\footnotesize Note: Training set contains 5,876 storm-related 
and 5,876 no-storm vortex regions from
120 days in 2008. 10-fold cross validation is performed in the evaluation. Testing set contains
2,706 storm-related cells and 7,773 no-storm cells from 60 days far away from the training data.
The feature vector for each vortex is composed by both visual features and storm priors (see
Section~\ref{sec:fea}). Beside the merged features (results shown in the first column),
we test the two types of features separately (results shown in the second and third columns).}
\end{table}

\begin{figure}
\includegraphics[width=\textwidth]{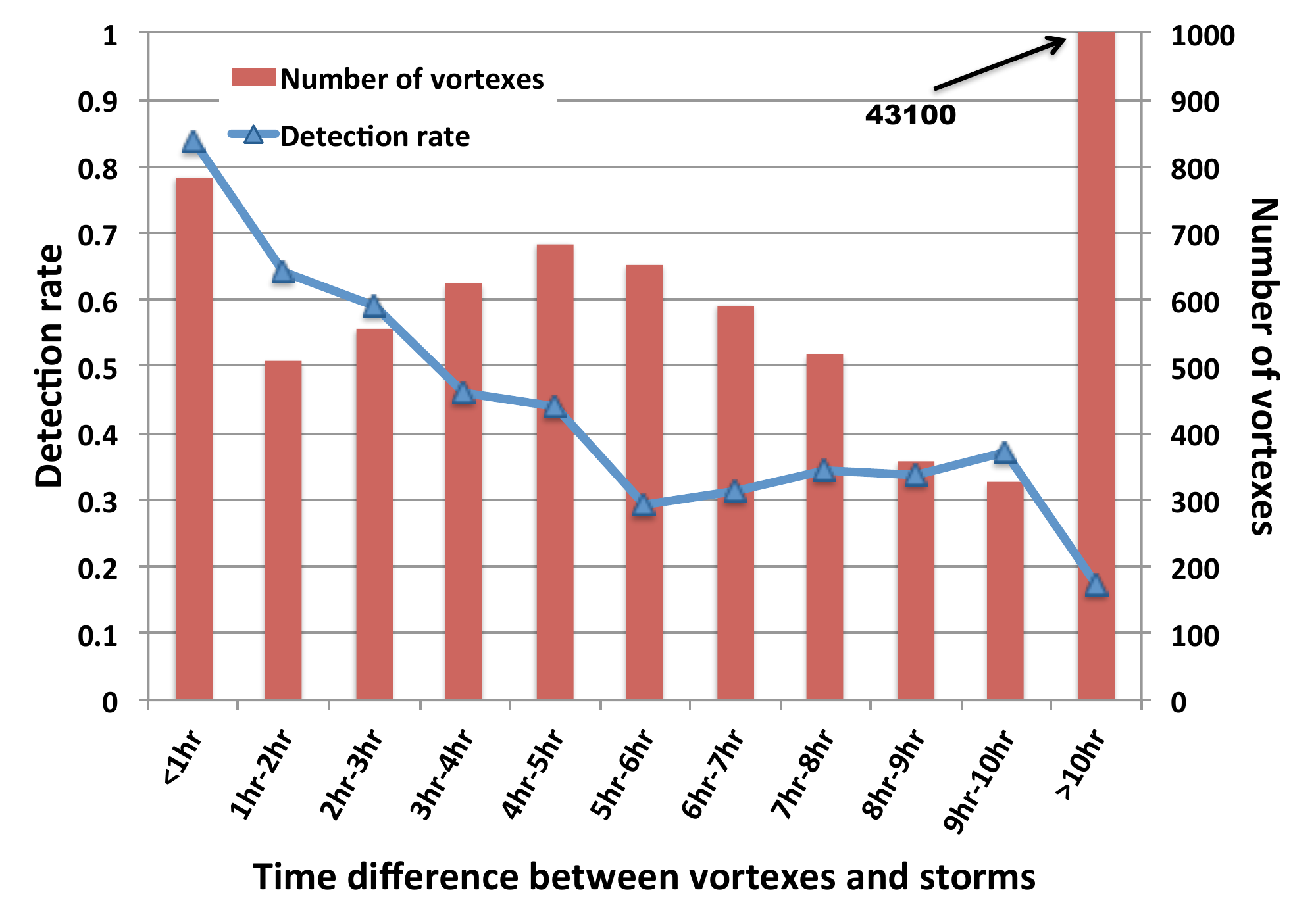}
\caption{Storm detection rate with different amounts of time in advance. The distribution
of the time differences ($\Delta t$) between the testing vortex samples and storms are plotted in red
bars (the last bar is not fully shown because it is much higher than others).
And the percentages of vortexes categorized as storms with different $\Delta t$ are
plotted in a blue curve. The figure shows that the predictivity of the classifier
is generally reliable within few hours of time in advance. The storm detection rate
decreases as the time difference becomes longer. }\label{fig:cls_pred}
\end{figure}

Secondly, we evaluated the classifier's ability to predict the future
locations of storms.
Instead of comparing each vortex's classification result with
its ground-truth label, which reflects its status within the period
of $(t-0.5hr, t+2hr)$ ($t$ is the timestamp of the vortex),
we observed how much time in advance a storm can be detected by our algorithm.
Given a testing vortex extracted at the location $(u,v)$ and time $t$,
we find the earliest time $t'=T(u,v,t)$
that a neighboring storm $(u',v',t') \in \mathscr{D}$
developed since two hours before $t$,
i.e.,
$$t'=\min_{t_i}{\left\{(u_i,v_i,t_i) \in \mathscr{D} \text{ and }\|u_i-u\| < 3^{\circ} \text{ and }\|v_i-v\| < 3^{\circ} \text{ and }t_i>t-2hr\right\}}\;.$$
The value $\Delta t = T(u,v,t)-t$ indicates the time difference between an observed vortex
and the first storm to form nearby storm. For a vortex with $\Delta t> 0$,
a future storm is predicted if
the classifier can label a vortex as storm-related. The percentages of storms
predicted for vortexes with different $\Delta t> 0$ are shown in Fig.~\ref{fig:cls_pred}.
The results show that the proposed algorithm can effectively predict future
storms within several hours of the detection of vortexes and 
the prediction reliability increases as $\Delta t$ decreases.
Particularly, it shows a reasonable prediction rate for the
development of nearby storms within a period of
4 hours after
the observation of visual vortexes in satellite images.

It is necessary to emphasize that the classification is performed
on already extracted candidate vortex regions where local clouds
are rotating. Most of the cloud-covered regions on the satellite
images without obvious rotational motions have been already eliminated
before the algorithm proceeds to the classification stage.
Therefore in practice the system achieves a good overall precision for
detecting the storms. In addition, the classification results for
individual vortex regions can be further corrected and validated based
on their spatial and temporal neighboring relationships.

\subsection{Case Study}
To demonstrate the usefulness of the proposed vortex detection and
classification algorithm in real scenarios, we present several case
studies where our algorithm is applied for automatic storm detection
from satellite images.

\begin{landscape}
\begin{figure*}
\centering
\subfigure[2008/05/25 12:15 GMT]{
    \includegraphics[width=0.25\linewidth]{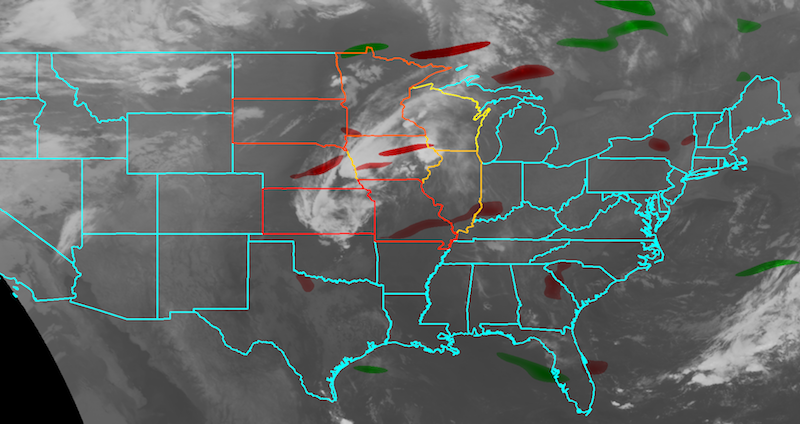}
}
\subfigure[2008/05/25 12:45 GMT]{
    \includegraphics[width=0.25\linewidth]{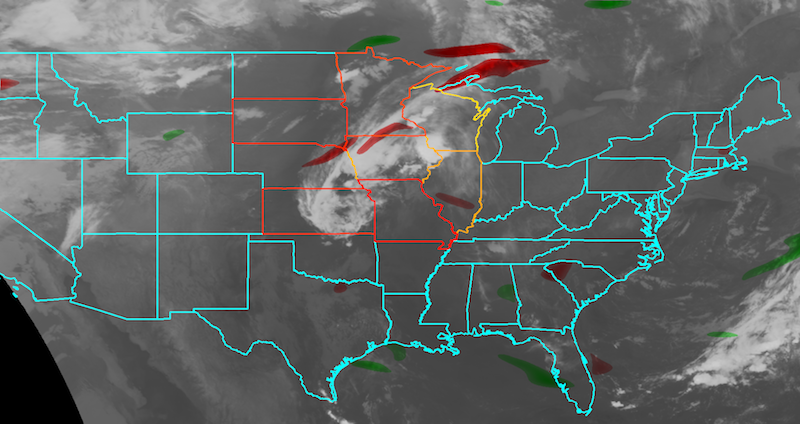}
}
\subfigure[2008/05/25 13:15 GMT]{
    \includegraphics[width=0.25\linewidth]{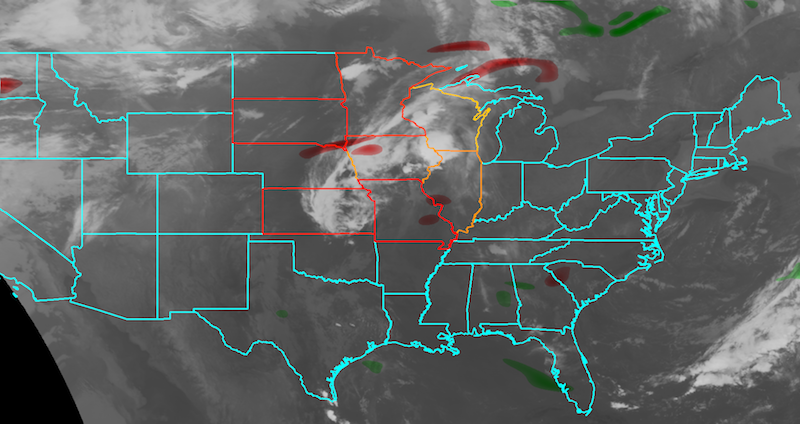}
}\\
\subfigure[2008/09/18 12:15 GMT]{
    \includegraphics[width=0.25\linewidth]{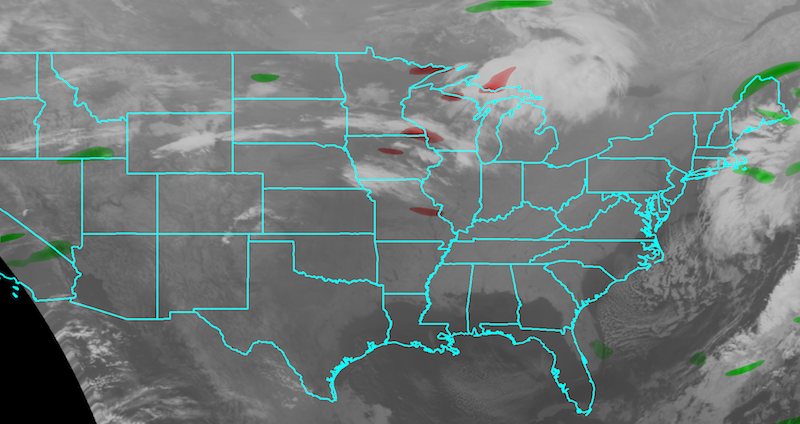}
}
\subfigure[2008/09/18 12:45 GMT]{
    \includegraphics[width=0.25\linewidth]{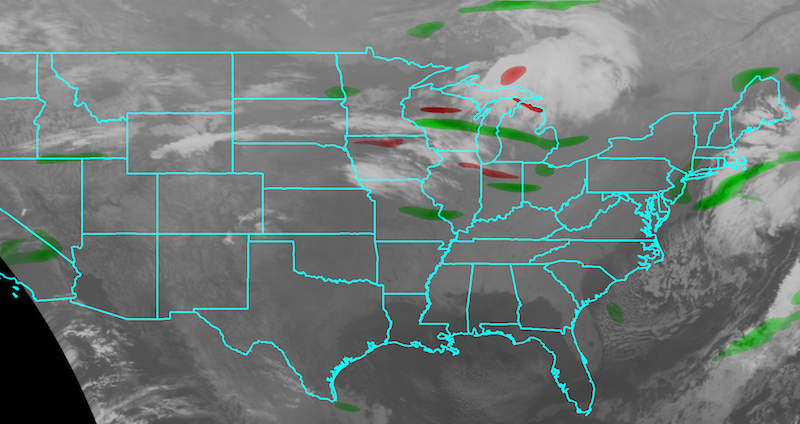}
}
\subfigure[2008/09/18 13:15 GMT]{
    \includegraphics[width=0.25\linewidth]{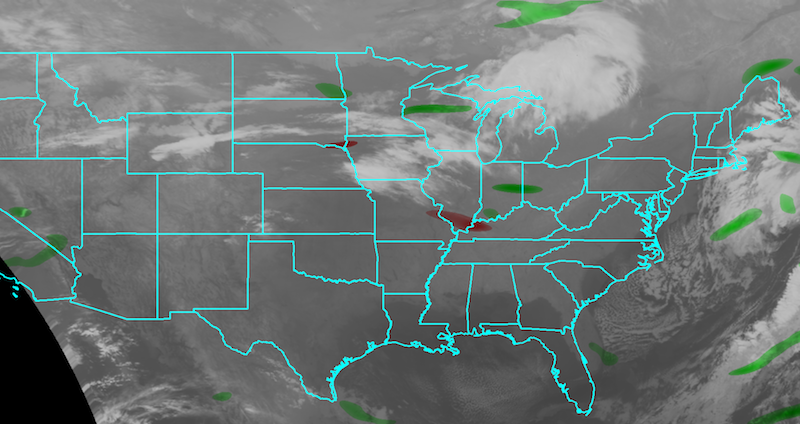}
}\\
\subfigure[2008/03/25 12:15 GMT]{
    \includegraphics[width=0.25\linewidth]{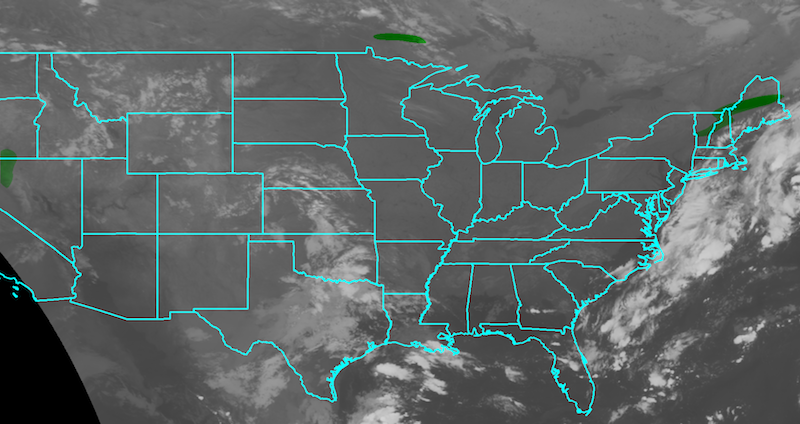}
}
\subfigure[2008/03/25 12:45 GMT]{
    \includegraphics[width=0.25\linewidth]{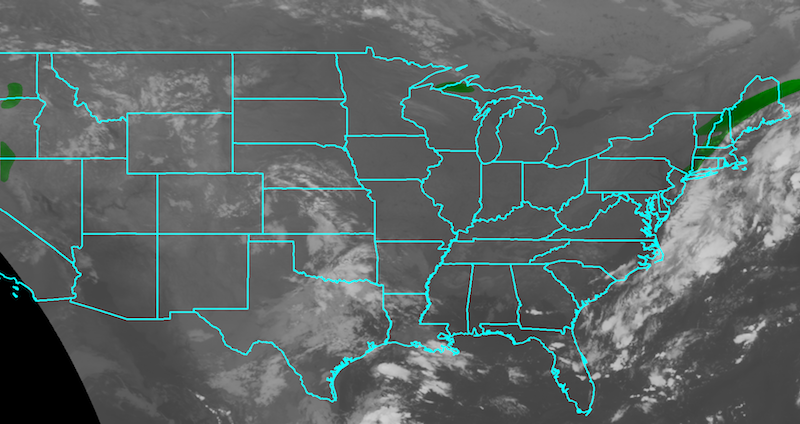}
}
\subfigure[2008/03/25 13:15 GMT]{
    \includegraphics[width=0.25\linewidth]{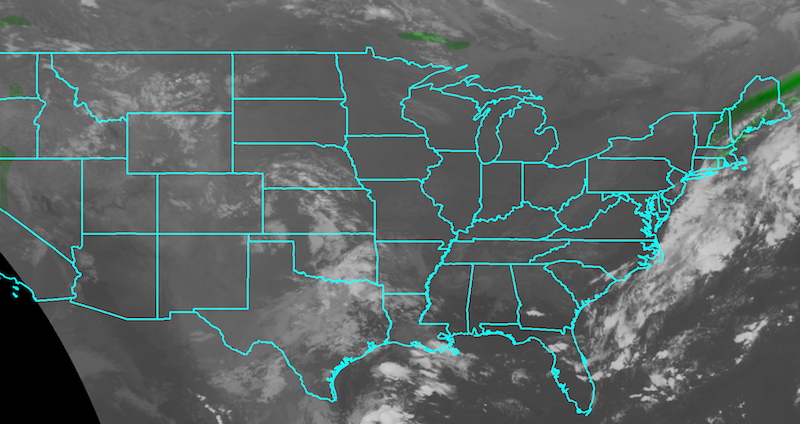}
}
\caption{Three example cases of applying the proposed algorithm to analyze
satellite image sequences.
Detected storm vortexes are colored in red and rejected vortexes are colored
in green. States hit by storms are highlighted in warm colors. A red state boundary
indicates the state is hit by an ongoing storm based on the historical report.
As the highlighting color turns to yellow, the corresponding state is affected by
more future storms as far as 6 hours away from the observation.
(a-c) Three frames on May 25, 2008. Major storms
occurred in the central, eastern, and southeastern states of US. Multiple
vortexes are detected and classified as storm systems in each frame and
the locations of the vortexes well match the ground-truth.
(d-f) Image frames on September 18, 2008. The US continent area is mostly
clear on the day and no vortex is detected. 
(g-i) Image frames on March 25, 2008. No storm is reported on the day.
Though clouds and their rotational motions are perceived on the image
sequence, most of them are categorized as hazardless
cloud systems by the classifier.
}\label{fig:case_study}
\end{figure*}
\end{landscape}

Three different cases are presented and the
results are visualized in Fig.~\ref{fig:case_study} (one row for each case).
An active storm reporting day , a clear storm-free day, and a cloudy storm-free day
are selected to demonstrate the adaptiveness of our algorithm.
These three image sequences are not included in the training process
of the storm classifier.
In Fig.~\ref{fig:case_study}, automatically detected vortexes are 
filled with colors on each image. Vortexes classified as storms
are colored in red, and the green vortexes are categorized
as storm-free cases by the machine learning module.
The US state boundaries are drawn on each image to show the
alignment between the satellite images and the geo-coordinate system.
In particular, boundaries of states affected by storms
are highlighted by warm colors from red to yellow. The highlighting
color indicates how long after the observation storms were reported
in the corresponding states. As the color turns to yellow, the highlighted
states are affected by storms further in the future.
Specifically, the red color means an ongoing storm and the yellow color
means a storm 6 hours after the observation time.
Fig.~\ref{fig:case_study} only shows the results on the first three
image frames of each case due to space limitation.
Each test image sequence in fact contains more than three frames,
and the storm detection results are consistent
in the subsequent frames for each case.



The first case is from May 25, 2008, which was an active storm day, when
several severe storms hit the Midwest US.
An obvious cloud vortex system is visible on the satellite
images and it is related to storms reported in nearby states.
Our algorithm captures all the major vortexes from the satellite
image sequence, and correctly categories most of them as storms.
The result also shows the capability of the proposed algorithm
to assist in short-term storm forecasting. In most cases there will
be storms detected around regions where storms were reported shortly thereafter.

Different from the first case, the second case is strongly negative
in that little clouds and vortexes can be perceived from the images.
Results on the second row of Fig.~\ref{fig:case_study} shows that
no storm is predicted because few clouds and vortexes can be captured
by the feature extraction stage of our algorithm. In this case, it
is easy for the algorithm to make correct decisions.



In the third case, several cloud systems can be observed above the
CONUS, and several vortexes are detected by the optical flow analysis.
In reality, even though some cloud vortex systems
look dominant on the image sequence,
there is no severe storm reported on that day.
Results shown on the third row of Fig.~\ref{fig:case_study}
demonstrate that the machine learning module successfully
avoids labeling the detected cloud vortexes as storms.
This shows that the system does not over-predict the occurrence of storms.
This benefit is also validated by the standalone test for the classifier
showing that specificity of the classifier is as high as its sensitivity.
\section{Conclusions and Future Work}\label{sec:conclusion}
In this paper, we present a thunderstorm 
detection algorithm that locates storm visual signatures
from satellite images.
The algorithm automatically analyzes the visual features
from satellite images and incorporates the historical meteorological records
to make storm predictions. As opposed to traditional ways of weather forecasting,
our approach primarily relies on the visual information from images and tries
to extract high-level storm signatures as meteorologists usually do manually.
Without using any sensory measurement (e.g., temperature, air
pressure, etc.) as numerical weather forecasting does, the algorithm
captures global cloud patterns solely from the satellite images, which
is helpful in the synoptic-scale storm prediction.
Additionally, the algorithm
takes advantage of big data, using historical storm reports
in the past years together with the satellite image archives
to learn the correspondence between visual
image signatures and the
occurrences of current and future storms. Experiments 
and case studies show that
the algorithm is effective and robust.

Properties of optical flows between every two images in a sequence
are the basic visual clues adopted in the work. 
The algorithm can estimate robust and smooth optical flows between two
images and determine the rotations in the flow field.
Unlike numerical weather forecasting methods that are sensitive to noise
and initial conditions, our approach can consistently detect
reliable visual storm clues hours before
the occurrence of a storm. Therefore, the results are useful for 
weather forecasts, especially storm forecasts.

The application of historical storm records and machine learning
boosts the performance of the storm extraction algorithm.
Standalone vortex detection from the optical flow is not sufficient
to make reliable predictions. The statistical model trained from
historical meteorological data together with the satellite image
visual features further selects the extracted vortexes and removes
vortexes not related to storms. In our current algorithm,
storm detection is based on individual vortex regions. In reality,
multiple vortexes tend to appear near each other in storm systems both temporally
and geographically. Therefore, taking into account nearby vortexes
within a frame and tracking storm systems across a sequence
can improve the overall reliability of the system. In particular,
tracking the development of a storm system will be helpful
for analyzing the future trend of the storm. This problem is nontrivial
because the storm systems (clouds) are non-rigid and highly variant,
and the same vortexes in a storm system are not always
detectable over time. Future work needs to tackle
the problem of non-rigid object tracking to better make use
of the temporal information in the satellite image sequences.

Lastly it should be emphasized that
weather systems are highly complex and chaotic, so it is always
a challenging task to make accurate weather forecasts. The proposed
algorithm based on the satellite image visual features
is not completely accurate and needs to be further validated.
We expect the algorithm or similar approaches
to play a part in producing weather forecasts
rather than making the forecasts alone.
It could be incorporated with other existing weather prediction
techniques (e.g. NWP) and used as an automatic tool to refine the
current forecasts.
It is useful and valuable because it
provides forecasts in a different aspect from the numerical approaches.
The purpose for developing the new algorithm is not to replace the
current weather forecasting models, but to produce additional
independent predictions to be combined with other approaches.
As a result, we will focus the future study on how to integrate
information from multiple data sources and predictions from different
models to produce more reliable and timely storm forecasts.



\ifCLASSOPTIONcompsoc
  \section*{Acknowledgments}
\else
  \section*{Acknowledgment}
\fi
This material is based upon work supported by the
National Science Foundation under Grant No. 1027854. Shared
computational infrastructure was provided by the Foundation under
Grant No. 0821527. Part of the work was done when J. Z. Wang and
J. Li were with the Foundation. Any opinions, findings, and
conclusions or recommendations expressed in this material are those of
the authors and do not necessarily reflect the views of the
Foundation. We also thank the US National Oceanic and Atmospheric
Administration (NOAA) for providing the data used in this
research. Siqiong He assisted with data collection. The discussions
with Jose A. Piedra-Fernandez of the University of Almeria, Spain, has
been very helpful.




\bibliographystyle{IEEEtran}
\bibliography{ref}



\end{document}